%% file: main.tex
\documentclass{article} 

\usepackage[accepted]{icml2021}

\usepackage{times}
\usepackage{multirow}
\usepackage{url}
\usepackage{qiangstyle}
\usepackage{mathrsfs}
\usepackage{amsmath}
\usepackage{tablefootnote}

\usepackage{multirow}
\usepackage{booktabs}

\usepackage{lipsum}  

\icmltitlerunning{AlphaNet: Improved Training of Supernets with Alpha-Divergence}

\renewcommand{\tt}{{p}}
\renewcommand{\ss}{{q}}
\newcommand{\amin}{\alpha_{-}}
\newcommand{\amax}{\alpha_{+}}
\renewcommand{\KD}{\mathrm{KD}}
\renewcommand{\L}{\mathcal{L}}
\newcommand{\qq}[1]{{\small\red{[QL:~#1]}}}

\newcommand{\erm}{\mathrm{ERM}}
\renewcommand{\D}{\mathrm{D}}

\begin{document}

\twocolumn[
\icmltitle{AlphaNet: Improved Training of Supernets with Alpha-Divergence}





\begin{icmlauthorlist}
\icmlauthor{Dilin Wang}{fb}
\icmlauthor{Chengyue Gong \textsuperscript{\tiny*}}{austin}
\icmlauthor{Meng Li \textsuperscript{\tiny *}}{fb}
\icmlauthor{Qiang Liu}{austin}
\icmlauthor{Vikas Chandra}{fb}
\end{icmlauthorlist}

\icmlaffiliation{austin}{Department of Computer Science, The University of Texas at Austin}
\icmlaffiliation{fb}{Facebook}

\icmlcorrespondingauthor{Dilin Wang}{wdilin@fb.com}
\icmlcorrespondingauthor{Chengyue Gong}{cygong@cs.utexas.edu}
\icmlcorrespondingauthor{Meng Li}{meng.li@fb.com}
\icmlcorrespondingauthor{Qiang Liu}{lqiang@cs.utexas.edu}
\icmlcorrespondingauthor{Vikas Chandra}{vchandra@fb.com}
\vskip 0.3in
]

\printAffiliationsAndNotice{\icmlEqualContribution}

\begin{abstract}

Weight-sharing neural architecture search (NAS) is an effective technique for automating efficient neural architecture design. 
Weight-sharing NAS builds a supernet that assembles all the architectures as its sub-networks and jointly trains the supernet with the sub-networks. The success of weight-sharing NAS heavily relies on distilling the knowledge of the supernet to the sub-networks. However, we find that the widely used distillation divergence, i.e., KL divergence, may lead to student sub-networks  that over-estimate or under-estimate the uncertainty of the teacher supernet, leading to inferior performance of the sub-networks. In this work, we propose to improve the supernet training with a more generalized $\alpha$-divergence.
By adaptively selecting the $\alpha$-divergence,
we simultaneously prevent the over-estimation or under-estimation of the uncertainty of the teacher model.  
We apply the proposed $\alpha$-divergence based supernets training to both slimmable neural networks and weight-sharing NAS, and demonstrate significant improvements. Specifically, our discovered model family, AlphaNet, outperforms prior-art models on a wide range of FLOPs regimes, including BigNAS, Once-for-All networks, and AttentiveNAS. We achieve ImageNet top-1 accuracy of 80.0\% with only 444M FLOPs.
Our code and pretrained models are available at \url{https://github.com/facebookresearch/AlphaNet}.

\end{abstract}

\input{tex/intro}

\input{tex/oneshot_nas}

\input{tex/adaptive_kd}

\input{tex/exp}

\input{tex/conclusion}

\bibliographystyle{iclr2019_conference}
\bibliography{supernet}

\input{tex/appendix}

\end{document}

%% file: tex/intro.tex
\section{Introduction}

Designing accurate and computationally efficient neural network architectures is an important but challenging task. Neural architecture search (NAS) automates the neural network design by exploring an enormous architecture space and achieves state-of-the-art (SOTA) performance on various applications including image classification~\citep{zoph2016neural, zoph2018learning}, object detection~\citep{ghiasi2019fpn}, semantic segmentation~\citep{zhang2019customizable} 

Conventional NAS approaches can be prohibitively expensive as hundreds of candidate architectures need to be trained from scratch and evaluated~\citep[e.g.,][]{tan2019mnasnet, zoph2018learning}. The supernet based approach has recently emerged to be a promising approach for efficient NAS. A supernet assembles all candidate architectures into a weight sharing network with each architecture corresponding to one sub-network. By training the sub-networks simultaneously with the supernet, different architectures can directly inherit the weights from the supernet for evaluation and deployment, which eliminates the huge cost of training or fine-tuning each architecture individually.

Though promising, simultaneously optimizing all sub-networks with weight-sharing is highly challenging for the supernet training~\citep[e.g.,][]{yu2020bignas, cai2019once}.
To stabilize the supernet training and improve the performance of sub-networks, one widely used approach is in-place knowledge distillation (KD) \citep{yu2019universally}. 
Inplace KD leverages the soft labels predicted by the largest sub-network in the supernet to supervise all the other sub-networks. By distilling the knowledge of the teacher model, the performance of the sub-networks can be improved significantly \citep{yu2019universally, yu2020bignas}. 

Standard knowledge distillation uses KL divergence to measure the discrepancy between the teacher and student networks. However, KL divergence penalizes the student model much more when it fails to cover one or more local modes of the teacher model~\citep{murphy2012machine}. Hence, the student model tends to over-estimate the uncertainty of the teacher model and suffers from inaccurate approximation of the most important mode, i.e., the correct prediction of the teacher model.

To further enhance the supernet training, we propose to replace the KL divergence with a more generalized $\alpha$-divergence~\citep{amari1985differential, minka2005divergence}. 
Specifically, by adaptively controlling $\alpha$ in the proposed divergence metric, we can penalize both the under-estimation and over-estimation of the teacher model uncertainty to encourage a more accurate approximation for the student models. 
While directly optimizing the proposed adaptive $\alpha$-divergence may suffer from a high variance of the gradients, we further propose a simple technique to clip the gradients of our adaptive $\alpha$-divergence to stabilize the training process.
We show the clipped gradients still define a valid divergence metric implicitly and hence, yielding a proper optimization objective for KD.

We empirically verify the proposed adaptive $\alpha$-divergence in two notable applications of supernets - slimmable networks \citep{yu2019universally} and weight-sharing NAS \citep{yu2020bignas, wang2020attentivenas} on ImageNet. For weight-sharing NAS, we train a supernet containing both small (200M FLOPs) and large (2G FLOPs) sub-networks following \citet{wang2020attentivenas}. With the proposed adaptive $\alpha$-divergence, we are able to train high-quality sub-networks, called AlphaNets, that surpass all prior state-of-the-art models in the range of 200 to 800 MFLOPs, like EfficientNets \citep{tan2019efficientnet}, OFANets~\citep{cai2019once}, and BigNas~\citep{yu2020bignas}. 
Specifically, AlphaNet-A4 achieves 80.0\% accuracy with only 444M FLOPs.

%% file: tex/oneshot_nas.tex
\section{Background}

Training high-quality supernets is fundamental for weight-sharing NAS but non-trivial~\citep{benyahia2019overcoming}. 
Recently, in-place KD is shown to be an effective mechanism that significantly improves the supernet performance \citep{yu2019universally, yu2020bignas}. 

To formalize the supernet training and in-place KD, consider a supernet with trainable parameter $\theta$.
Let $\mathcal A$ denote the collection of all sub-networks contained in the supernet. 
The goal of training a supernet is to learn $\theta$ such that all the sub-networks in $\mathcal A$ can be optimized simultaneously to achieve good accuracy.

\begin{figure}[tb]
\centering
\begin{tabular}{c}
\includegraphics[width=0.44\textwidth]{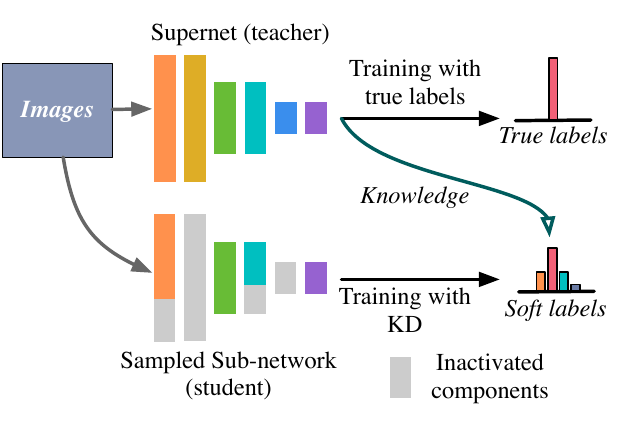} \\
\end{tabular}
\caption{An illustration of training supernets with KD. Sub-networks are part of the supernet with weight-sharing. }
\label{fig:supernet_kd}
\end{figure}

The supernet training process with the in-place KD is illustrated in Figure~\ref{fig:supernet_kd}. At each training step, given a mini-batch of data, the supernet as well as several sub-networks are sampled. While the supernet is trained with the real labels, all the sampled sub-networks are supervised with the soft labels predicted by the supernet. Then, the gradients from all the sampled networks are aggregated before the supernet parameters are updated. More formally, at the training step $t$, the supernet parameters $\theta$ are updated by
$$
\theta_{t} \gets \theta_{t - 1} + \epsilon g(\theta_{t - 1}),
$$
where $\epsilon$ is the step size, and
\begin{equation}
    \label{eq:iterative-update}
    g(\theta_{t - 1}) = 
    \nabla_{\theta} \bigg  (\L_{\mathcal D}(\theta) +
    \gamma 
    \E_{s} \L_{\KD}([\theta,s]; \theta_{t - 1}) \bigg )\bigg|_{\theta=\theta_{t - 1}}. 
\end{equation}
Here, $\L_{\mathcal D}(\theta)$ is the standard cross entropy loss of the supernet on a training dataset $\mathcal D$, $\gamma$ is the weight coefficient, and $\L_{KD}([\theta, s];~ \theta_{t})$ is the KD loss for distilling the supernet into a randomly sampled sub-network $s$, for which KL divergence has been widely used~\citep[e.g.,][]{yu2020bignas}.

Let $p(x; \theta)$ and $q(x; \theta, s)$ denote the output probability of the supernet and the sub-network $s$ given input $x$, then, we have
\begin{equation}
    \label{eq:kdloss}
    \!\L_{\KD}([\theta, s], \theta_t) 
    = 
    \E_{x\sim \mathcal D}
    [\KL (p(x;\theta_t) ~||~ q(x;\theta,s))],
\end{equation}
where $\KL(\tt~||~\ss)=\E_{\tt}[\log (\tt/\ss)]$. Note that the gradient on $p(x; \theta_t)$ in the KD loss is stopped as~\eqref{eq:kdloss} indicated. For notation simplicity, 
we denote $p$ as our teacher model and $q$ (or $q_\theta$) as student models in the following. 

Additionally, note that the way KD is used in the supernet training is different from the standard settings such as \citet[e.g.,][]{hinton2015distilling}, where the teacher network is pre-trained and fixed.

%% file: tex/adaptive_kd.tex
\section{Supernet training with $\pmb{\alpha}$-divergence}

In this section, we analyze the limitations of using KL divergence in KD and propose to replace KL divergence with a more generalized $\alpha$-divergence. We study the impact of different choices of $\alpha$ values in the proposed divergence metric and further propose an adaptive algorithm to select $\alpha$ values during the supernet training. Meanwhile, we also show that while directly optimizing $\alpha$-divergence is challenging due to large gradient variances, a simple clipping strategy on $\alpha$-divergence can be very effective to stabilize the training.

\begin{figure*}[t]
\centering
\setlength{\tabcolsep}{6pt}
\begin{tabular}{ccc}
\raisebox{2.5em}{\rotatebox{90}{\small Prediction}}
\includegraphics[height=0.20\textwidth]{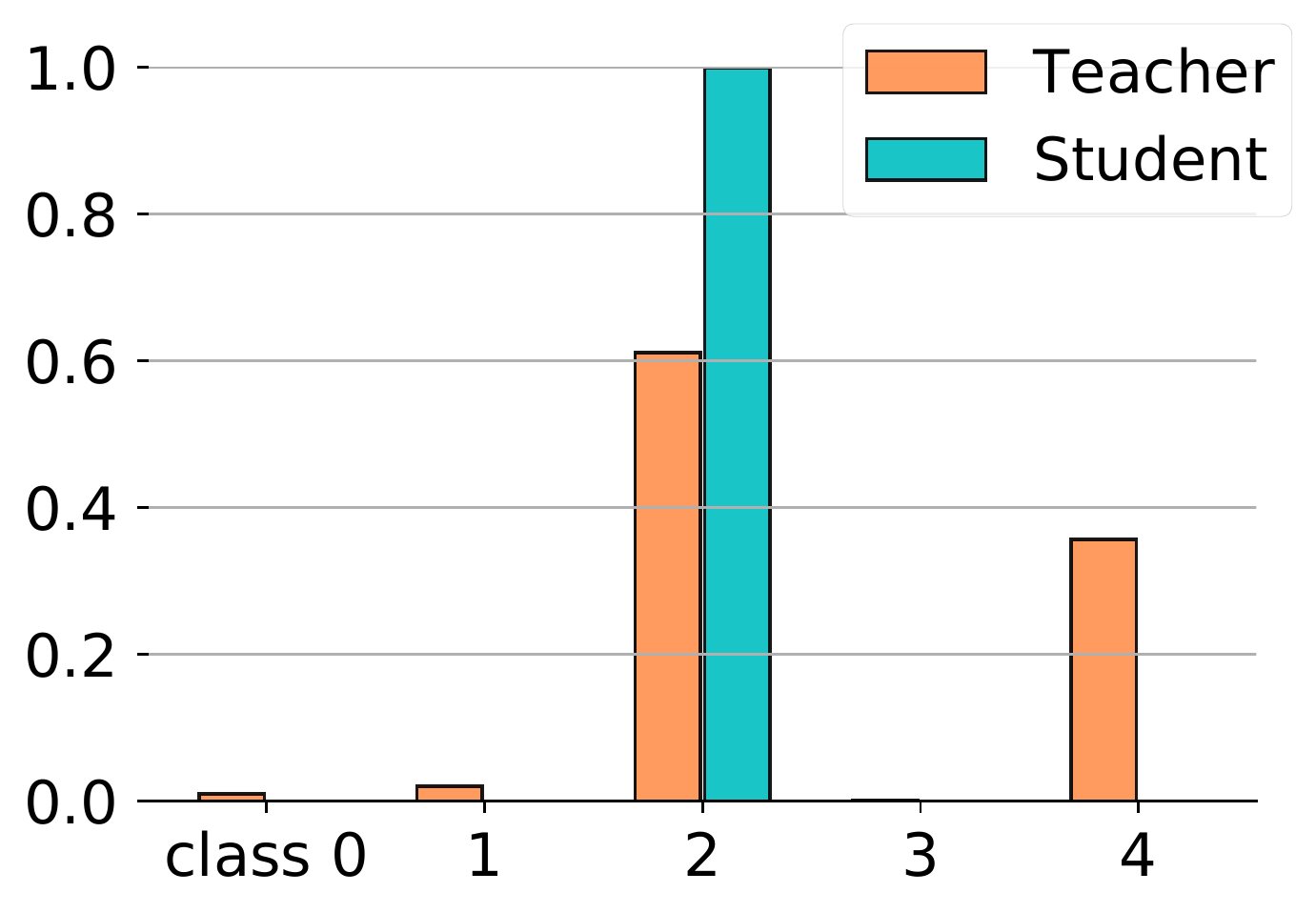} & 
\raisebox{2.5em}{\rotatebox{90}{\small Prediction}}
\includegraphics[height=0.20\textwidth]{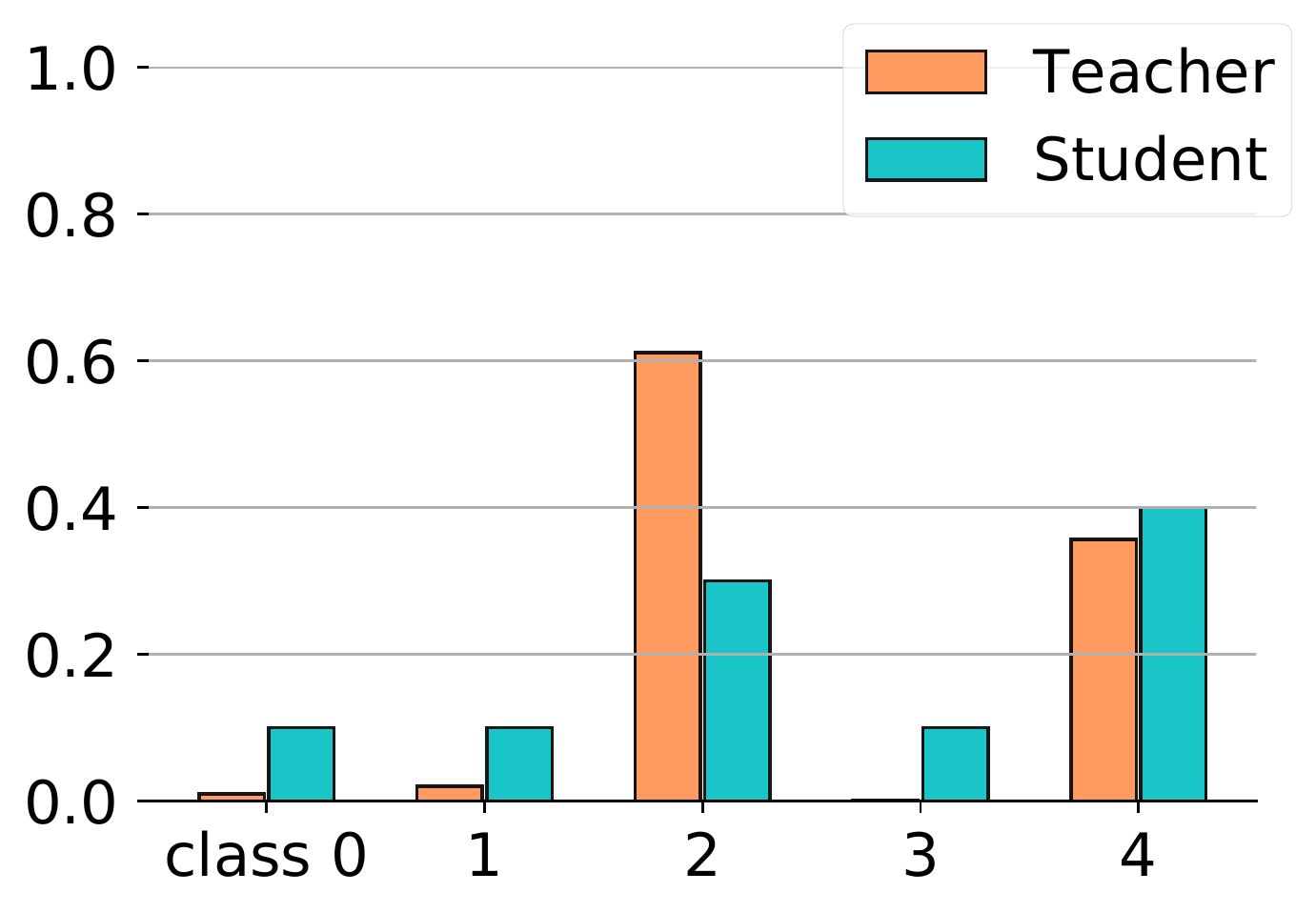} & 
\raisebox{2.0em}{\rotatebox{90}{\small $\alpha$-divergence}}
\includegraphics[height=0.20\textwidth]{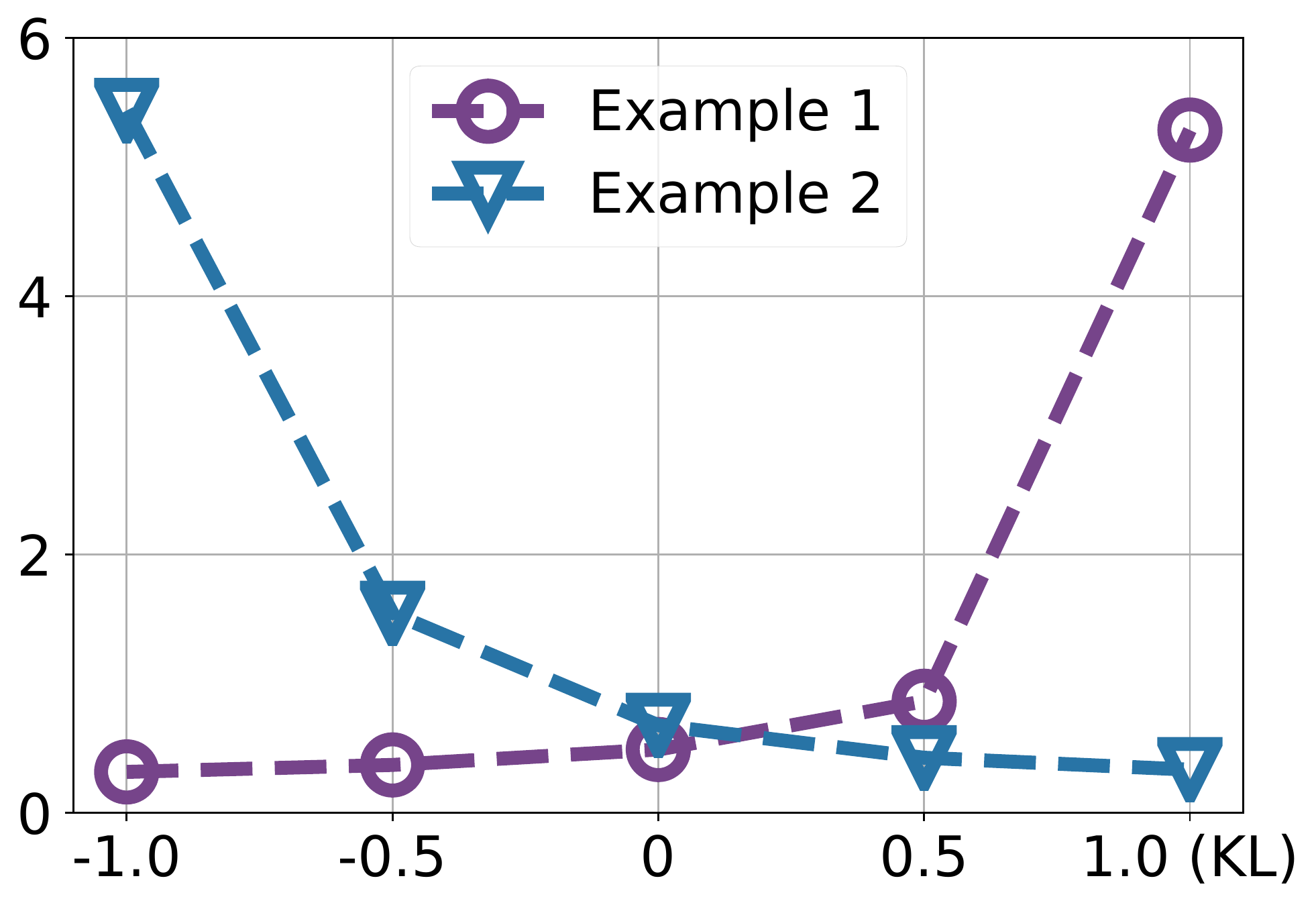} \\ 
\small (a) Example 1  \emph{(under-estimation)}  & \small  (b) Example 2 \emph{(over-estimation)} &\small  (c) choices of $\alpha$ \\
\end{tabular}
\caption{
(a) \emph{Example 1 - uncertainty under-estimation}.
The student network under-estimates the uncertainty of the teacher model and misses important local modes of the teacher model. 
 (b) \emph{Example 2 - Uncertainty over-estimation}. In this case, 
the student network over-estimates the uncertainty of the teacher model and misclassifies the most dominant mode of the teacher model. (c) plots the corresponding $\alpha$-divergences between the student model and the teacher model for \emph{Examples 1} and \emph{2}. Note that $\KL$ divergence is a special case of $\alpha$-divergences with $\alpha=1$. We refer to the uncertainty as the entropy of predictions after the Softmax layer of the network.}
\label{fig:overview_alpha}
\end{figure*}

\subsection{Classic KL based KD and its limitations}
\label{sec:alpha_div}

KL divergence has been widely used to measure the discrepancy in output probabilities between the teacher and student models in KD. 
One main drawback with KL divergence is that it cannot sufficiently penalize the student model when it over-estimates the uncertainty of the teach model.
Let $\tt$ and $\ss$ denote the output probability of the teacher and student models, respectively. The KL divergence between the teacher and student models is calculated by $\KL(\tt||\ss) = \E_{\tt}[\log (\tt/\ss)]$. When $\tt > 0$, to ensure $\KL(\tt || \ss)$ remains finite, we must have $\ss > 0$. This is the so-called \emph{zero avoiding} property of $\KL$. In contrast, when $\tt = 0$, $\ss > 0$ does not get penalized. 
For example, as shown in Figure \ref{fig:overview_alpha} (b) and (c), even though the student model over-estimates the uncertainty of the teacher model and predicts the wrong class ("class 4"), the KL divergence is still small.

The aforementioned \emph{over-estimation} in Example 2 would be penalized
at a larger magnitude when using other types of divergences, e.g., reverse KL divergence $\KL(\ss || \tt)$. 
For reverse KL divergence, 
$\KL(\ss~||~\tt) = \E_{\ss}[\log (\ss/\tt)]$ is infinite if $\tt = 0$ and $\ss>0$. 
Hence if $\tt =0$ we must ensure $\ss=0$, this is known as the \emph{zero forcing} property~\citep{murphy2012machine}. 
Therefore, minimizing reverse KL divergence encourages the student model $\ss$ to avoid low probability modes of $\tt$ while focusing on the modes with high probabilities, and thus, may \emph{under-estimate} the uncertainty of the teacher model, as shown in Example 1 in Figure~\ref{fig:overview_alpha}. 

Hence, a natural question is whether it is possible to generalize the KL divergence to simultaneously suppress both the under-estimation and over-estimation of the teacher model uncertainty during the supernet training.

\subsection{KD with adaptive $\pmb{\alpha}$-divergence}
Our observations shown in Figure~\ref{fig:overview_alpha} motivate us to design a new KD objective that 
simultaneously penalize both over-estimation and under-estimation of the teacher model uncertainty. 
We first generalize the typical $\KL$ divergence with a more flexible $\alpha$-divergence~\citep{minka2005divergence}. 
Consider $\alpha \in \RR \setminus \{0,1\}$, 
the $\alpha$-divergence is defined as 
\begin{align}
D_\alpha(\tt~||~\ss) = \frac{1}{\alpha(\alpha-1)} \sum_{i=1}^m q_i \bigg[ \bigg( \frac{\tt_i}{\ss_i} \bigg)^\alpha - 1 \bigg],  
\label{equ:alpha_divergence}
\end{align}
where $q=[q_i]_{i=1}^m$ and $p=[p_i]_{i=1}^m$ are two discrete distributions on $m$ categories. 
The $\alpha$-divergence includes a large spectrum of classic divergence measures. In particular, 
the KL divergence  ${\KL}(\tt ~||~ \ss)$  is the limit of $D_\alpha(\tt ~||~ \ss)$ with $\alpha \to 1$ while the reverse KL divergence ${\KL}(\ss ~||~ \tt)$ is the limit of $D_\alpha(\tt ~||~ \ss)$ with $\alpha \to 0$.

A key feature of $\alpha$-divergence is that we can decide to focus on penalizing different types of discrepancies (under-estimation or over-estimation) by choosing different $\alpha$ values. 
For example, 
as shown in Figure \ref{fig:overview_alpha} (c), 
when $\alpha$ is negative, 
$D_\alpha(\tt~||~\ss)$ is large when $\ss$ is more widely spread than $\tt$
(when $\ss$ \emph{over-estimates} the uncertainty in $\tt$), 
and is small when $\ss$ is more concentrated than $\tt$ 
(when $\ss$ \emph{under-estimates} the uncertainty in $\tt$). 
The trend is opposite when $\alpha$ is positive:

To simultaneously alleviate the over-estimation and under-estimation problem when training the supernet, we consider a positive $\amax$ together with a negative $\amin$, and propose to
use the maximum of $D_{\amax}(p~||~q)$
and $D_{\amin}(p~||~q)$ in the KD loss function: 
$$
D_{\amax,\amin}(\tt~\pp~\ss) = \max\bigg\{ \underbrace{D_{\amin}(\tt~\pp~\ss)}_\text{\scriptsize 
\begin{tabular}{c}penalizing\\ over-estimation\end{tabular}}, ~\underbrace{D_{\amax}(\tt~\pp~\ss)}_\text{\scriptsize \begin{tabular}{c}penalizing\\ under-estimation\end{tabular} } \bigg\}.
$$

Our KL loss now changes from  Eqn.~\eqref{eq:kdloss} to 
\begin{equation}
    \label{eq:adaptive-alpha-divergence}
        \!\!\L_{KD}([\theta, s], \theta_t) 
        = 
        \E_{x\sim \mathcal D}
        [D_{\amax,\amin} (p(x;\theta_t)~||~ q(x;\theta,s))].
\end{equation}
We denote this KD strategy that always chooses the maximum of $D_{\amin}$ and $D_{\amax}$ to optimize as \emph{Adaptive-KD}.

\subsection{Stabilizing $\pmb{\alpha}$-divergence KD}
\label{sec:stablize}

One would prefer to set both $|\amax|$ and $|\amin|$ to be large to ensure the student model is sufficiently penalized when it either under-estimates or over-estimates the uncertainty the teacher model.
However, 
directly optimizing the $\alpha$-divergence with large $|\alpha|$ is often challenging in practice. 
Consider the gradient of $\alpha$-divergence:
$$
    \nabla_\theta D_\alpha (p ~||~q_\theta) = -\frac{1}{\alpha}\E_{\ss_\theta}\bigg[   \bigg(\frac{\tt}{\ss_\theta}\bigg)^\alpha  \nabla_\theta \log \ss_\theta \bigg]. 
$$
If $|\alpha|$ is large, then the powered term $({\tt}/{\ss_\theta})^\alpha$ can be quite significant and cause the training process to be unstable.
To enhance the training stability, we clamp the maximum value of $({\tt}/{\ss_\theta})^\alpha$ to be $\beta$, and obtain 
\begin{align}
\tilde \nabla_\theta D_\alpha(p~||~q_\theta) \overset{\mathrm{def}}{=}  -\frac{1}{\alpha}\E_{\ss_\theta}\bigg[  \mathrm{Clip}_\beta\bigg( \frac{\tt}{\ss_\theta}\bigg)^\alpha  \nabla_\theta \log \ss_\theta \bigg],
\label{eq:grad_stablize_alpha_div}
\end{align}
where
 $\mathrm{Clip}_\beta(t) = \min(t, \beta)$. 

Eqn.~\eqref{eq:grad_stablize_alpha_div} is a simple yet effective heuristic approximation of $\nabla_\theta D_{\alpha}(p~||~q_\theta)$. It is important to note that Eqn.~\eqref{eq:grad_stablize_alpha_div} equals the \emph{exact} gradient of a special $f$ divergence between $p$ and $q_\theta$. Hence, our updates still amount to minimizing a valid divergence. 
Note that the clipping function $\mathrm{Clip}_\beta(\cdot)$ is only partially differentiable. 
So naively clipping on $(\tt/\ss_\theta)^\alpha$ in Eqn.~\eqref{equ:alpha_divergence} may stop gradients back-propagating from the density ratio terms, 
hence yielding gradients that are not from a valid divergence. 

To show that we still optimize a valid divergence with Eqn.~\eqref{eq:grad_stablize_alpha_div}, note that, for a convex function $f \colon [0, +\infty)\to \RR$, 
the $f$-divergence between $p$ and $q_\theta$ is defined as 
\begin{align}
D_f(p ~||~q_\theta) = \E_{\ss_\theta}\bigg[  f\bigg(\frac{\tt}{\ss_\theta} \bigg) - f(1)\bigg].
\end{align}
Its gradient w.r.t. $\theta$ is 
\begin{align*}
    \nabla_\theta D_f(\tt~||~\ss_\theta) 
     = -\E_{\ss_\theta} \bigg[ \rho_f\bigg(\frac{\tt}{\ss_\theta}\bigg) \nabla_\theta \log \ss_\theta  \bigg],
\end{align*}
where $\rho_f(t) = f'(t)t - f(t)$ (\citet{wang2018variational}). 
Note that $\alpha$-divergence is a special case of $f$-divergence when 
$f(t) = t^\alpha / (\alpha(\alpha-1))$. 

\begin{pro}
There exists a convex function $f\colon (0,+\infty)\to \RR$, such that 
$\tilde \nabla_\theta D_\alpha (p~||~q_\theta)$ in~\eqref{eq:grad_stablize_alpha_div} is the exact gradient of $D_f(p~||~q_\theta),$ that is, 
$\tilde \nabla_\theta D_\alpha (p~||~q_\theta) = \nabla_\theta D_f(p~||~q_\theta)$. 
\end{pro}
\begin{proof}
Let $\rho_*(t) = \frac{1}{\alpha} \mathrm{Clip}_\beta(t)^\alpha$. 
We just need to find a $f$ such that 
$$\rho_f(t) = f'(t) t - f(t) = \rho_*(t). 
$$
Taking derivation on both sides, we get 
$
f''(t)t = \rho_*'(t). 
$
This gives $f''(t) = \rho_*'(t)/t$ 
and hence $f(t) = \iint \rho_*'(t)/t dt$,
where  $\iint$ denotes second-order antiderivative (or indefinite integral). 
Because $\rho_*(t)$ is non-decreasing, 
we have $\rho_*'(t)/t \ge 0$ for $t>0$, and hence $f$ is convex on $(0,+\infty)$. 
\end{proof}

In practice, we apply Eqn.~\eqref{eq:grad_stablize_alpha_div} to the $\alpha$-divergence used in Eqn.~\eqref{eq:adaptive-alpha-divergence}.
By clipping the value of importance weights, what we optimize is still a divergence metric but is more friendly to gradient-based optimization.

%% file: tex/exp.tex
\begin{table*}[ht]
\centering
\setlength{\tabcolsep}{4pt}
\begin{tabular}{c|l|ccccccccccc}
\hline
Model & Method & 0.25$\times$ &  0.3$\times$  &  0.35$\times$  &  0.4$\times$  &  0.45$\times$  & 0.5$\times$  &  0.55$\times$   &  0.6$\times$  & 0.65$\times$ & 0.7$\times$ & 0.75$\times$\\
\hline
\hline
\multirow{3}{*}{MbV1} & w/o KD & 53.9 & 55.3 & 57.1 & 59.1 & 61.1 & 62.9 & 64.0 & 65.8 & 66.9 & 67.9 & 68.8 \\  
&w/ KL-KD  &  \bf 56.4  & 57.8 & 59.5 & 61.0 & 63.0 & 64.4 & 65.5 & 67.1 & 68.3 & 69.1 & 69.8 \\
& \bf w/ Adaptive-KD  (ours) & \bf 56.4 & \bf 57.9 & \bf  59.7 &  \bf 61.7 & \bf  63.4 &  \bf 65.0 &  \bf 66.2 &  \bf 67.7 &  \bf 68.8 &  \bf 69.5 &  \bf 70.1\\ \hline \hline 
\multirow{3}{*}{MbV2} &  w/o KD & - & - & 61.9 & 62.8 & 63.7 & 64.5 & 65.1 & 67.2 & 67.7 & 68.3 & 69.0 \\
& w/ KL-KD  &  - & - &63.2 & 64.4 & 65.1 & 66.0 & 66.5 & 68.4 & 69.2 & 69.5 & 70.1  \\
& \bf  w/ Adaptive-KD  (ours) &  - & - & \bf 63.7 & \bf64.6 &\bf 65.6 & \bf66.3 &\bf 66.9 & \bf68.7 &\bf 69.3 &\bf 69.9 & \bf70.5 \\ \hline 
\end{tabular}
\caption{Top-1 validation accuracy on ImageNet for Slimmable MobileNetV1 networks (denoted by MbV1) and Slimmable MobileNetV2 networks (denoted by MbV2) 
trained with different KD strategies.}
\label{tab:slimmbale}
\end{table*}

\begin{table*}[ht]
\centering
\setlength{\tabcolsep}{4pt}
\begin{tabular}{c|l|ccccccccccc}
\hline
Model & Method & 0.25$\times$ &  0.3$\times$  &  0.35$\times$  &  0.4$\times$  &  0.45$\times$  &  0.5$\times$  &  0.55$\times$   &  0.6$\times$  & 0.65$\times$ & 0.7$\times$ & 0.75$\times$\\
\hline \hline 
\multirow{3}{*}{MbV1} & w/ KL-KD (T=0.5)~~~~
& 55.1  & 56.0 & 57.6& 59.1  & 61.4 & 62.5  & 64.0 & 65.6  & 66.9  & 67.9  & 68.7  \\
& T=2.0 & 55.4& 57.0 & 58.8 & 60.7 & 62.6 & 64.1& 65.3 & 66.6  & 67.9 & 68.7& 69.5\\
&  T=4.0 & 48.7  & 50.7  & 53.1 & 55.9  & 58.8 & 60.9& 62.7 & 64.6 & 66.0 & 67.4 & 68.3\\
\hline \hline 
\multirow{3}{*}{MbV2} & w/ KL-KD (T=0.5)~~~& - & - &  61.7 & 62.9 & 63.8 & 64.6 & 65.0 & 67.4 & 68.4 & 68.8 & 69.8  \\
 & T=2.0 & - & - & 62.6 & 63.9 & 64.8 & 65.6 & 66.4 & 68.1 & 68.6 & 69.1 & 70.0 \\ 
 & T=4.0 & - & - & 59.3 & 60.9 & 62.2 & 63.1 & 64.0 & 66.3 & 67.1 & 67.7 & 68.8 \\ \hline 
\end{tabular}
\caption{Comparison to KL based KD with different temperature (T).  
We report top-1 validation accuracy on ImageNet for slimmable MobileNetV1 and MobileNetV2  networks, denoted by MbV1 and MbV2, respectively.}
\label{tab:slimmable_abalation}
\end{table*}

\section{Experiments}
We apply our \emph{Adaptive-KD} to improve notable supernet-based applications, including  slimmable neural networks \citep{yu2019universally}
and weight-sharing NAS  \citep[e.g.,][]{cai2019once, yu2020bignas, wang2020attentivenas}. 
We provide an overview of our algorithm for training the supernet in Algorithm~\ref{alg:main}.

\begin{algorithm}[t]
\caption{Training supernets with $\alpha$-divergence}
\label{alg:main}
\begin{algorithmic}[1]
\STATE {\bf Input}: Adaptive $\alpha$-divergence range given by $\amin$ and $\amax$, a clipping factor $\beta$, a supernet with parameter $\theta$, and a search space $\mathcal A$.
\WHILE{not converging}
\STATE Sample a mini-batch of data $B$.
\STATE Train the supernet with true labels from $B$
\STATE Draw $k$ subnetworks $\{s_1, \cdots, s_k\} $ from $\mathcal A$; train sub-networks to mimic the supernet on the mini-batch data $B$ with the KD loss defined in Eqn.~\eqref{eq:adaptive-alpha-divergence} using clipped gradients in Eqn.~\eqref{eq:grad_stablize_alpha_div}.
\ENDWHILE
\end{algorithmic}
\end{algorithm}

\paragraph{Adaptive-KD settings}
In our algorithm, $\amin$ and $\amax$ control the magnitude of penalizing on \emph{over-estimation} and \emph{under-estimation}, respectively. And, $\beta$ controls the range of density ratios between the teacher model and the student model.
We find our method performs robustly w.r.t. a wide of range of choices of $\amin, \amax$ and $\beta$, yielding consistent improvements over the KL based KD baseline.  
Throughout the experimental section, we set $\amin=-1$, $\amax=1$ and $\beta=5.0$ as default for our method. We provide detailed ablation studies on these hyper-parameters in section~\ref{sec:exp_ablation}. 

\subsection{Slimmable  Neural Networks}
\label{sec:exp_slimmable}
Slimmable neural networks~\citep{yu2018slimmable, yu2019universally} are examples of supernets that support a wide range of channel width configurations. The search space $\mathcal{A}$ of slimmable networks contains networks with different width and all the other architecture configurations (e.g. depth, convolution type, kernel size) are the same.
This way, 
slimmable networks allow different devices or applications to adaptively adjust the model width on the fly according to on-device resource constraints to achieve the optimal accuracy vs. energy efficiency trade-off.

\paragraph{Settings}
We closely follow the training recipe provided in \citet{yu2019universally}, 
and use slimmable MobileNetV1 \citep{howard2017mobilenets} 
and slimmable MobileNetV2 \citep{sandler2018mobilenetv2} as our testbed.
Specifically, we train slimmable MobileNetV1 to support arbitrary dynamic width in the range of $[0.25, 1.0]$, and train slimmable MobileNetV2 to support 
dynamic widths of $[0.35, 1.0]$. 

We adopt the sandwich rule sampling proposed in~\citet{yu2019universally} for training. At each training iteration, 
we sample the largest sub-network with the largest channel width, 
the smallest sub-network with the smallest channel width and two random sub-networks to accumulate the gradients.
We train the supernet with ground truth labels and 
train all subsampled sub-networks with KD following~\eqref{eq:iterative-update}. 
For our baseline KD strategy, we set the KD coefficient $\gamma$ to be the number of sub-networks sampled, i.e., $\gamma=3$, as default  following~\citet{yu2019universally}.
To evaluate the effectiveness of our method, 
we simply replace the baseline KL-based KD loss used in \citet{yu2019universally} with our adaptive KD loss in~\eqref{eq:adaptive-alpha-divergence}. 

Additionally, we train all models for 360 epochs using SGD optimizer with momentum as 0.9, weight decay as $10^{-5}$ and dropout as 0.2. 
We use cosine learning rate decay, with an initial learning rate of 0.8, and batch size of 2048 on 16 GPUs.  
Following~\citet{yu2019universally}, we evaluate on ImageNet~\citep{deng2009imagenet}. 
We note that the baseline models trained with our hyper-parameter settings 
outperform those reported in \citet{yu2019universally}.

\paragraph{Results} 
We summarize our results in Table~\ref{tab:slimmbale}. 
We report the top-1 accuracy on the ImageNet.
Here, \emph{w/o KD} denotes the training strategy that excludes the effect of KD.  All such sub-networks are trained with ground truth labels via cross entropy. 

As we can see from Table~\ref{tab:slimmbale}, 
both baseline KL based KD (denoted as \emph{w/ KL-KD}) and our \emph{adaptive KD} (denoted as \emph{w/ Adaptive-KD}) yield significant performance improvements compared to \emph{w/o KD}. 
Our results confirm the importance of KD for training Slimmable networks. 
Meanwhile, our Adaptive-KD further improves on KL based KD for all the channel width configurations evaluated for both Slimmable MobileNetV1 (denoted by MbV1) and Slimmable MobileNetV2 (denoted by MbV2).

\begin{figure*}[t]
\centering
\setlength{\tabcolsep}{2pt}
\begin{tabular}{ccc}
\raisebox{1.5em}{\rotatebox{90}{\small Top-1 validation accuracy}}
\includegraphics[width=0.3\textwidth]{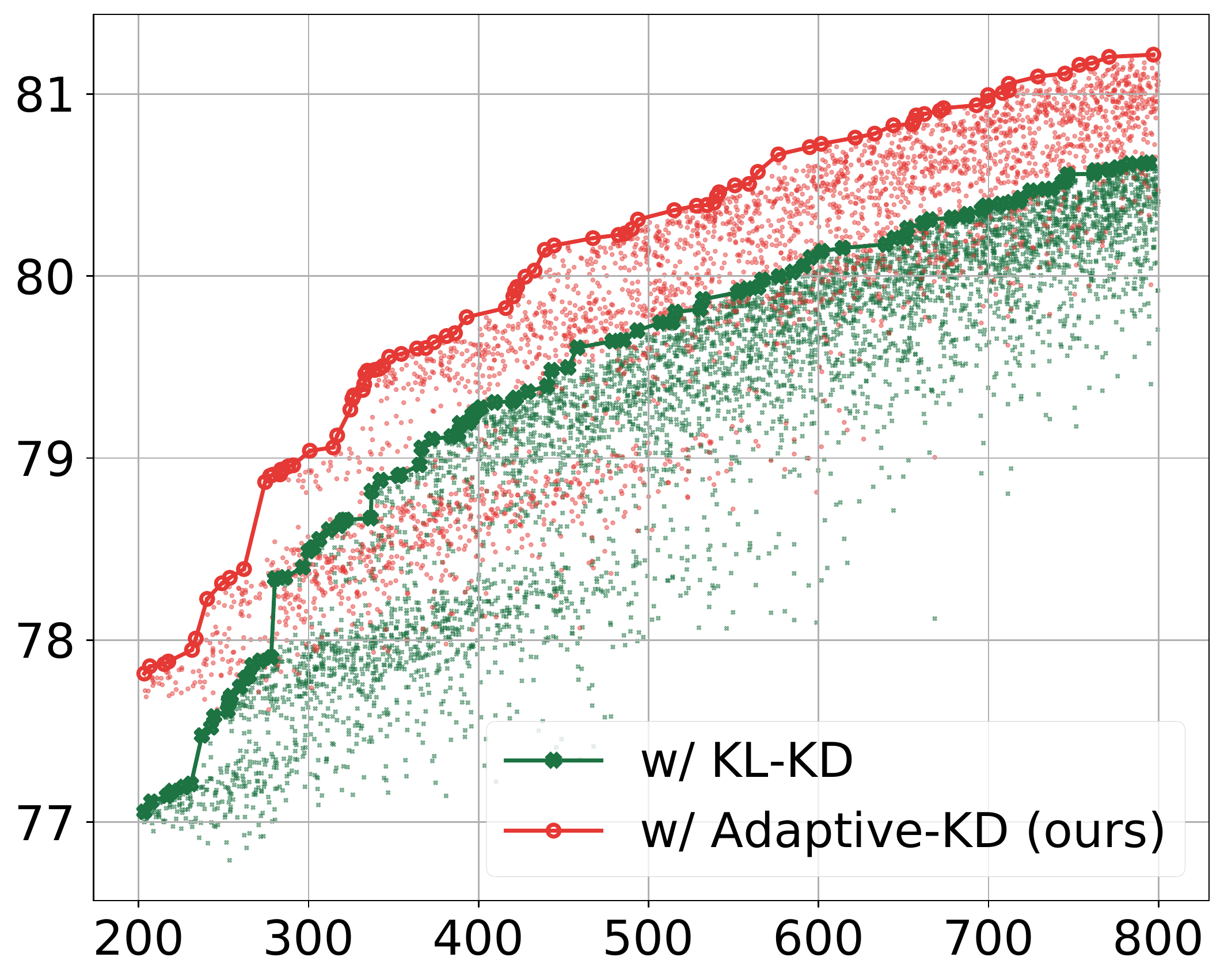} &
\includegraphics[width=0.3\textwidth]{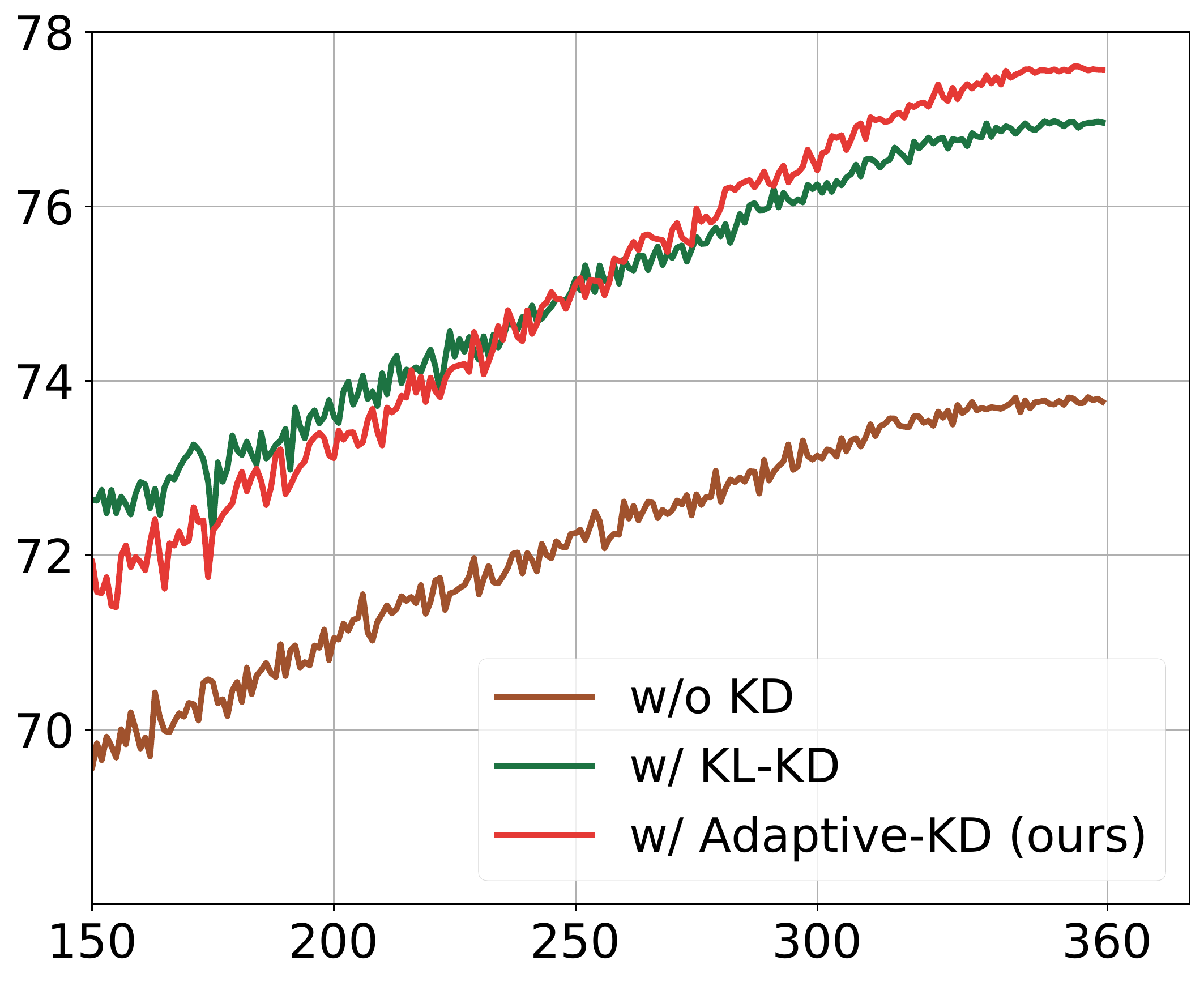} &
\includegraphics[width=0.3\textwidth]{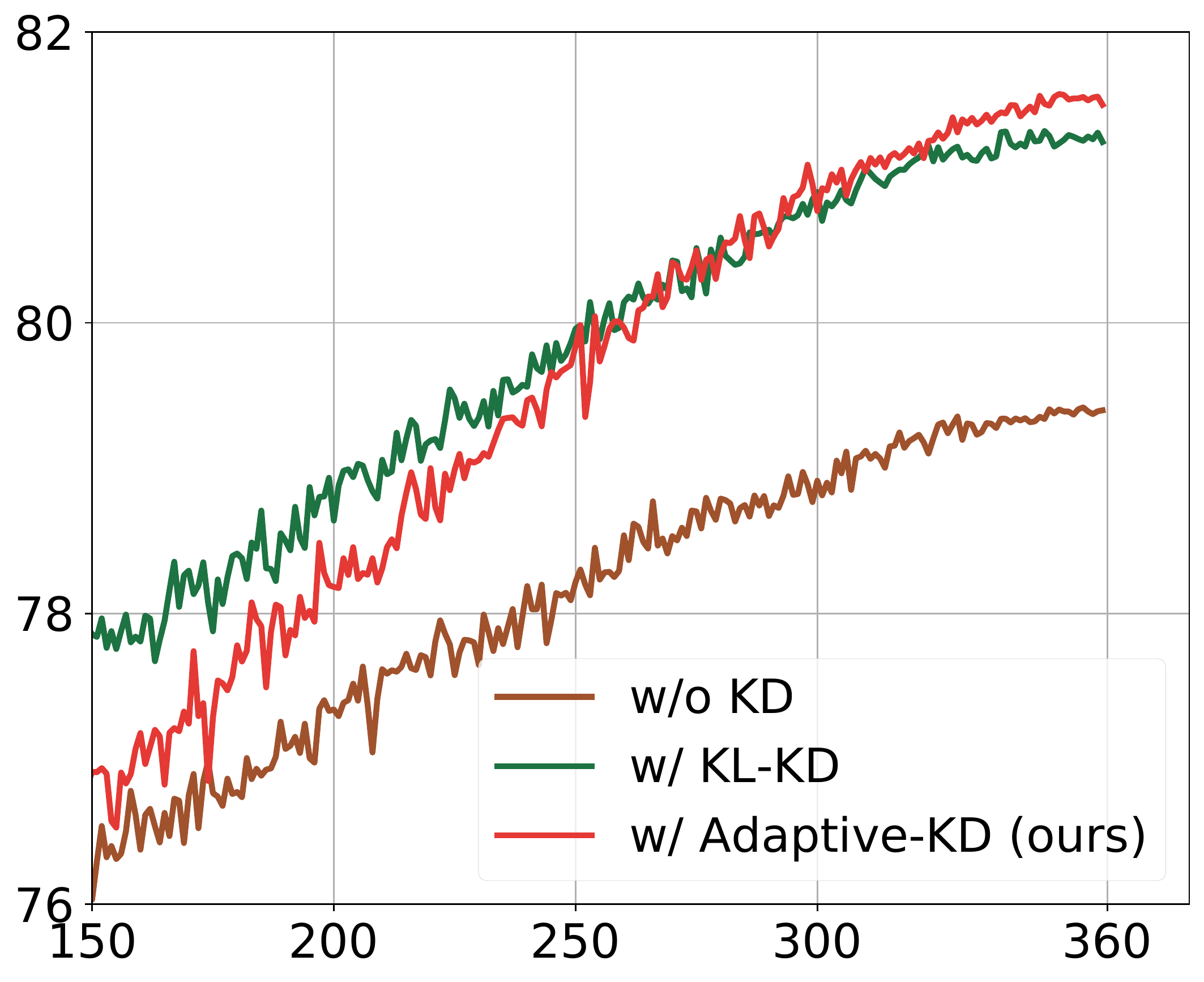} \\
\small  M FLOPs & \small Training Epoch & \small  Training Epoch \\
\small (a) Performance Pareto front & \small (b) Training curve of the smallest sub-network & \small (c) Training curve of the supernet \\ 
\vspace{-1.0em}
\end{tabular}
\caption{(a) Comparison of Pareto-set performance of the supernet trained via KL based KD and our adaptive KD, respectively. Each dot represents a sub-network evaluated during the evolutionary search step. 
(b-c) Training curves of the smallest sub-network and the largest sub-network (i.e., the supernet). }
\vspace{-1em}
\label{fig:supernet_pareto_and_training}
\end{figure*}

\begin{figure*}[t]
\vspace{1em}
\begin{tabular}{c}
\raisebox{3em}{\rotatebox{90}{\small Validation Accuracy}}
\includegraphics[width=0.93\textwidth]{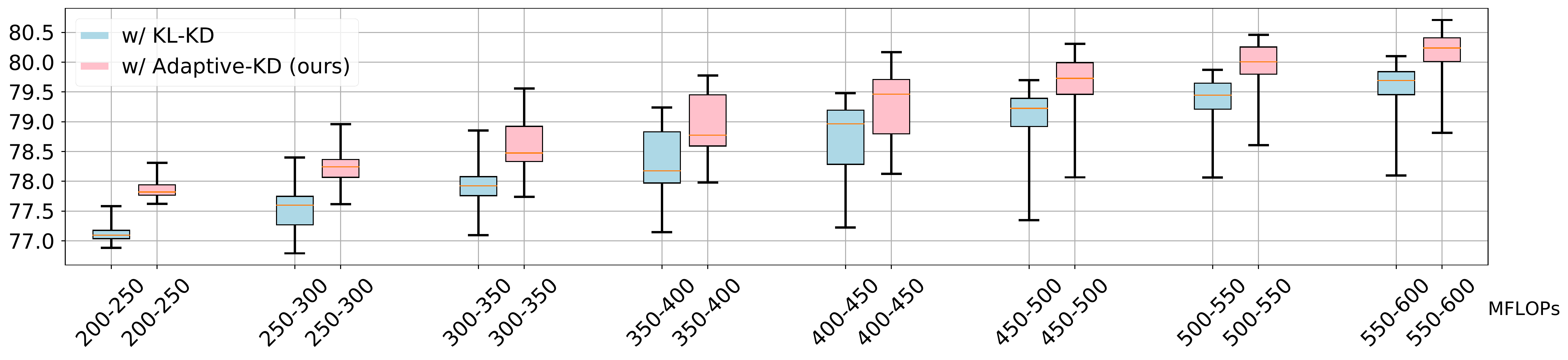}  \\
\end{tabular}
\caption{Top-1 accuracy on ImageNet from weight-sharing NAS with KL-based KD and adaptive-KD. Each box plot shows the performance of sampled sub-networks within each FLOPs regime. From bottom to top, each horizontal bar represents the minimum accuracy, the first quartile, the median, the third quartile and the maximum accuracy, respectively. }
\label{fig:supernet_boxplot}
\end{figure*}

\paragraph{Comparison to KD with different temperature coefficients}
As discussed in~\citet{hinton2015distilling}, 
for standard KL based KD, one can soften (or sharpen) the probabilities of the teacher and the student model by applying a temperature in their softmax layers. The best distillation performance might be achieved with a different temperature other than the normally used temperature of $1$. 

To ensure a fair comparison,  
we further evaluate the baseline KL based KD under different temperature ($T$) settings following the approach in~\citet{hinton2015distilling}. We refer the reader to  Appendix~\ref{app:kd} for detailed discussion on this topic.
In particular, we test a number of temperatures - 0.5, 2 and 4. 
We summarize our results in Table~\ref{tab:slimmable_abalation}.
We find all these settings to systematically perform worse 
than the simple KD strategy without temperature scaling, i.e., $T=1$. 
Additionally, the models trained via our method yield the best performance.

\begin{table*}[ht]
    \centering
    \setlength{\tabcolsep}{3pt}
    \begin{tabular}{l|ccccccc}
    \hline 
     & A0 {\scriptsize (203M)}   & A1{\scriptsize (279M)}  & A2{\scriptsize (317M)}  & A3{\scriptsize (357M)} & A4{\scriptsize (444M)}  & A5 {\scriptsize (491M)} & A6 {\scriptsize (709M)}   \\ \hline 
    w/o KD & 73.8 & 75.4 & 75.6 & 76.0 & 76.8 &77.1 &  77.9  \\
    w/ KL-KD & 77.0 & 78.2 & 78.5 &78.8 &79.3 &79.6 & 80.1 \\
    w/ Symmetric KL-KD & 77.0 & 78.4 &  78.5 & 78.7 & 79.3 & 79.5 &  79.9\\
    w/ KL-KD + Attentive Sampling \textsuperscript{$\dagger$} & 77.3 & 78.4 & 78.8 & 79.1 & 79.8 & 80.1 & 80.7 \\ \hline 
    \bf  w/ Adaptive-KD (ours - AlphaNet) &   \bf77.8 & \bf 78.9  & \bf79.1 & \bf 79.4 &\bf 80.0 & \bf 80.3 & \bf 80.8  \\
    \hline  
    \end{tabular}
    \caption{
    Performance on the discovered networks in \citet{wang2020attentivenas}. 
    Each (\#M) denotes the FLOPs of the corresponding model. 
    \textsuperscript{$\dagger$} uses additional attentive sampling \citep{wang2020attentivenas} for training the supernet.
    We denote our models as AlphaNet models. Here symmetric KL refers to a combination of
    the $\KL$ and the reverse $KL$ divergence, i.e., $\KL(q~||~p)+\KL(p~||~q)$.
    }
    \label{tab:attnas}
\end{table*}

\subsection{Weight-sharing NAS}
\label{sec:exp_supernet}
We apply our \emph{Adaptive-KD} to improve the training of the supernet for weight-sharing NAS~\citep{cai2019once, yu2020bignas, wang2020attentivenas}. Please see Appendix~\ref{app:one_shot_nas} for a brief introduction on weight-sharing NAS. 
Note that one main procedure of weight-sharing NAS is to simultaneously train all sub-networks specified in the search space to convergence. 
Similar to training Slimmable neural networks, 
this is often achieved by enforcing all sub-networks to learn from the supernet with KL based KD, ~\citep[e.g.,][]{yu2020bignas}. 

\paragraph{Training.} 
Our training recipe follows~\citet{wang2020attentivenas} except we use uniform sampling for simplicity. We pursue minimum code modifications to ablate the effectiveness of our KD strategy. 
We evaluate on the ImageNet dataset~\citep{deng2009imagenet}. 
All training details and the search space we used are discussed in Appendix~\ref{app:supernet}.

We use the update rule defined in \eqref{eq:iterative-update} to train the supernet.
Following~\citet{wang2020attentivenas} and \citet{yu2020bignas}, 
at each iteration, 
we train the supernet with ground truth labels and simultaneously 
we train the smallest sub-network and two random sub-networks with KD. 
In this way, a total of $4$ networks are trained at each iteration.

\begin{figure}[h]
\centering
\begin{tabular}{c}
\raisebox{2.5em}{\rotatebox{90}{Top-1 validation accuracy}}
\includegraphics[width=0.42\textwidth]{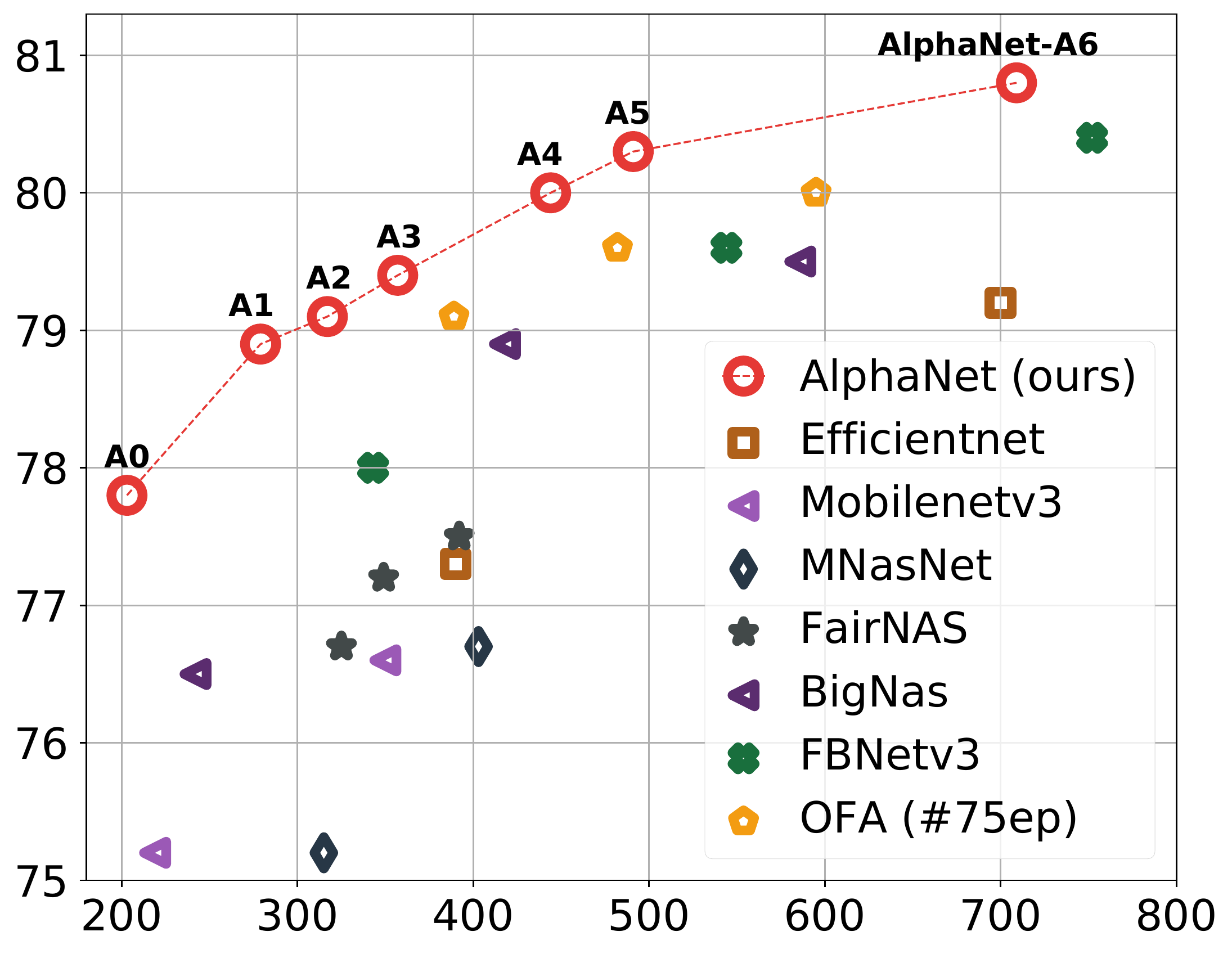}
\\
\small M FLOPs\\
\end{tabular}
\caption{Comparison with prior art NAS approaches on ImageNet. 
\#75ep denotes the models are further finetuned for 75 epochs with weights inherited from the corresponding supernet.}
\label{fig:compare_with_sota}
\end{figure}

\paragraph{Evaluation}
We compare the accuracy vs. FLOPs Pareto formed by the supernet learned by different KD strategies. 
To estimate the performance Pareto, we proceed as follows: 
1) we first randomly sample 512 sub-networks from the supernet and estimate their accuracy on the ImageNet validation set;  
2) we apply crossover and random mutation on the best performing 128 sub-networks following~\citet{wang2020attentivenas}. 
We fix both the crossover size and mutation size to be 128, yielding 256 new sub-networks. We then evaluate the performance of these sub-networks; 
3) We repeat the second step 20 times. 
The total number of sub-networks thus evaluated is $5,376$.

\paragraph{Results}
As we can see from Figure~\ref{fig:supernet_pareto_and_training}(a), 
\emph{Adaptive-KD} achieves a significantly better Pareto frontier compared to the KL-based KD baseline (denoted as \emph{w/ KL-KD}) and 
the simple training strategy without KD (denoted as \emph{w/o KD}).
Figures~\ref{fig:supernet_pareto_and_training}(b) and (c) plot the convergence curve of the smallest sub-network and the supernet, respectively. 
Our method adaptively optimizes a more difficult KD loss between the supernet and the sub-networks, yielding slightly slower convergence in the early stage of the training but better performance towards the end of the training. 

In Figure~\ref{fig:supernet_boxplot}, we group sub-networks according to their FLOPs and visualize five statistics for each group of sub-networks, including the minimum, the first quantile, the median, the third quantile and the maximum accuracy. Our method learns significantly better sub-networks in a quantitative way.

\paragraph{Improvement on SOTA}
As we use the same search space as in~\citet{wang2020attentivenas},
we further evaluate the discovered AttentiveNAS models (from A0 to A6)
with the supernet weights learned by our adaptive KD.
We refer to our models as AlphaNet models. 

As we can see from  Table~\ref{tab:attnas}, 
our Adaptive-KD significantly improves on classic KL based KD, yielding 
an average of $0.7\%$ improvements in the top-1 accuracy from A0 to A6. 
We aslo compare with symmetric KL based KD (namely, $\KL(p || q) +
KL(q || p)$). The corresponding results are no
better than those by using standard KL based KD training.
This is probably because the two different KL terms produce
conflicted gradients during training, which may therefore
lead to inferior final performance.
Additionally, our AlphaNet outperform all corresponding AttentiveNAS models~\citep{wang2020attentivenas}, which requires building Pareto-aware sampling distributions with additional computational overhead.

We further compare our AlphaNet against prior art NAS baselines, including  EfficientNet~\citep{tan2019efficientnet}, FBNetV3~\citep{dai2020fbnetv3}, BigNAS~\citep{yu2020bignas}, OFA~\citep{cai2019once}, MobileNetV3~\citep{howard2019searching}, FairNAS~\citep{chu2019fairnas} and  MNasNet~\citep{tan2019mnasnet}, in Figure~\ref{fig:compare_with_sota}. 
Our method outperforms all the baselines evaluated, establishing new SOTA accuracy vs. FLOPs trade-offs on ImageNet. 
For example, our model achieves 77.8\% top-1 accuracy with only 203M FLOPs. Under similar FLOPs constraint, the  corresponding top-1 accuracy is 75.2\%  with 219M FLOPs for MobileNetV3, 76.5\% top-1 accuracy with 242M FLOPs for BigNAS. 
Compared to OFA, our model achieves the same 80.0\% top-1 accuracy with 35\% fewer FLOPs (444M \emph{v.s.} 595M) and the same 79.1\% top-1 accuracy with  26\% fewer FLOPs (317M \emph{v.s.} 400M).

\begin{table}[t]
    \centering
    \setlength{\tabcolsep}{2pt}
    \begin{tabular}{l|cc|cc}
    \hline 
    Dataset &  Eff-B0 & Alp-A0 & Eff-B1 & Alp-A6 \\ \hline 
    Oxford Flowers & 97.2 & \bf{97.7} & 97.8 & \bf 98.7 \\
    Oxford-IIIT Pets& 91.2 & \bf{91.5} & 92.4 & \bf 92.9\\
    Food-101 & 87.6  & \bf{88.3} & 89.0 & \bf 89.6\\
    Stanford Cars & 91.0& \bf{91.5} & 92.2 & \bf 92.6\\
    FGVC Aircraft & 88.1  & \bf{88.5} & 88.7 & \bf 89.1\\
    \hline 
    \end{tabular}
    \caption{
    Comparison of transfer learning accuracy.
    `Eff' and `Alp' denotes EfficientNet and AlphaNet, respectively.
    All the networks are pretrained on ImageNet and then finetuned on transfer learning datasets. 
    EfficientNet-B0 and B1 has a model size of 390 MFLOPs and 700 MFLOPs, respectively. AlphaNet-A0 and A6 use 203 MFLOPs  and 709 MFLOPs, respectively. 
    }
    \vspace{-10pt}
    \label{tab:transfer_learning}
\end{table}

\begin{figure*}[ht]
\centering
\begin{tabular}{cc}
\raisebox{2.5em}{\rotatebox{90}{\small Relative Accuracy}}
\includegraphics[height=0.27\textwidth]{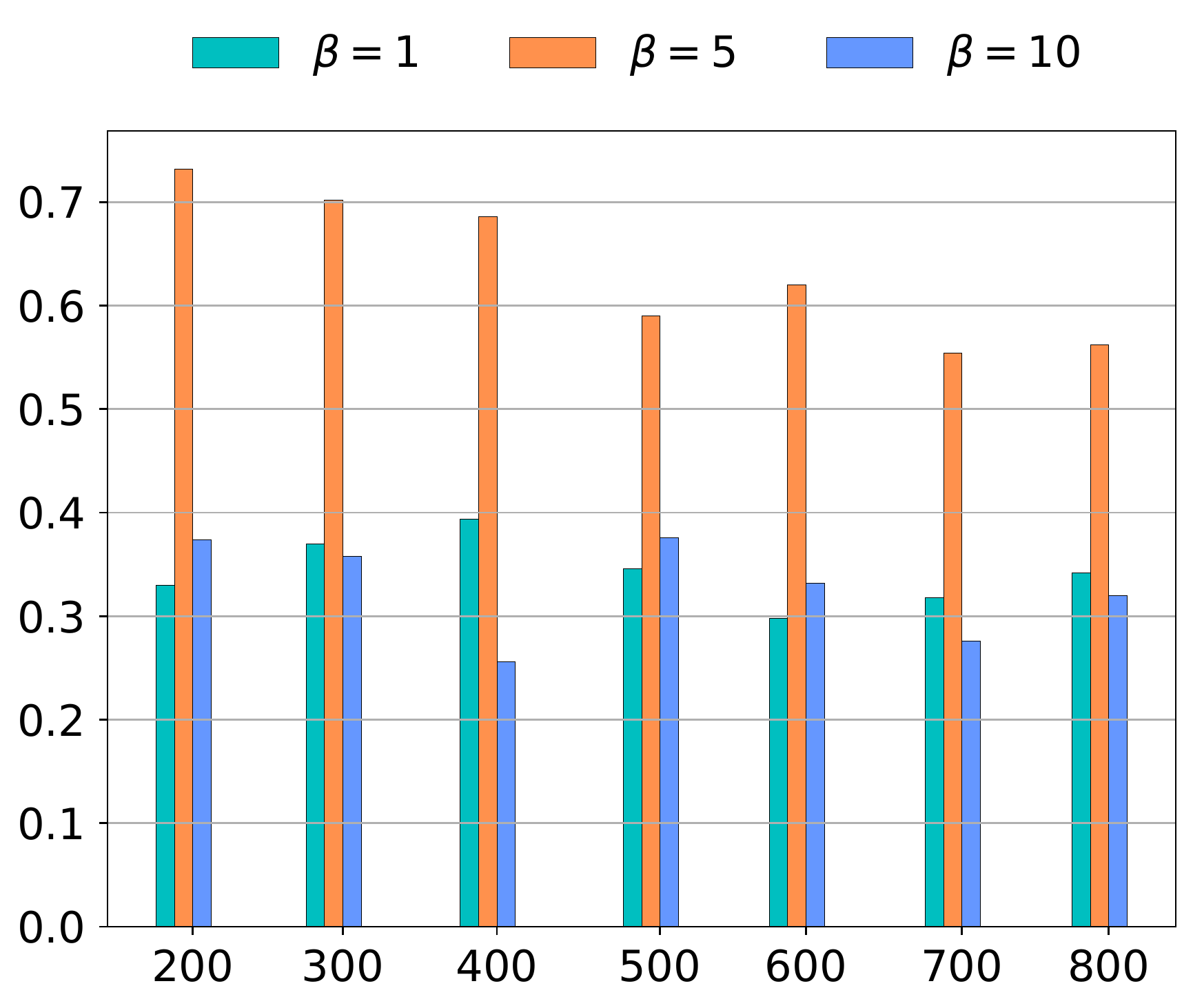} &
\raisebox{2.5em}{\rotatebox{90}{\small Relative Accuracy}}
\includegraphics[height=0.27\textwidth]{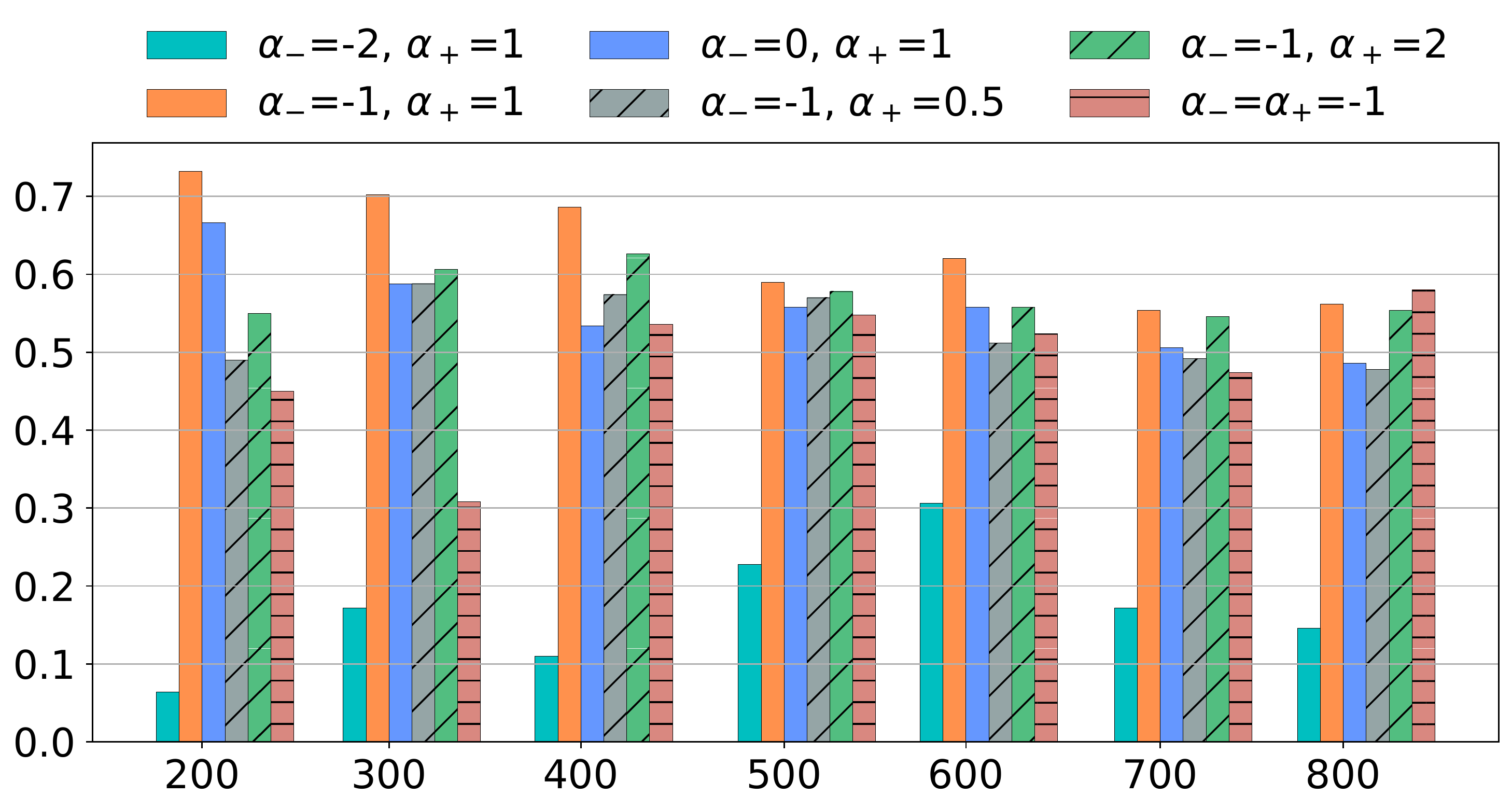} \\
\small MFLOPs $\pm50$ & \small MFLOPs $\pm50$\\
\small (a) Ablation study of $\beta$ & \small (b) Ablation study of $\amin$ and $\amax$\\
\end{tabular}
\vspace{-0.5em}
\caption{Relative accuracy compared to the results of KL based KD. 
Figure (a): we fix $\amin=-1, \amax=1$ and study the effect of our clipping factor $\beta$. 
Figure (b): we set $\beta=5$ as default and study the impact of $\amin$ and $\amax$. 
}
\label{fig:abla_alpha}
\end{figure*}

\begin{table*}[ht]
    \centering
    \begin{tabular}{l|c|ccc|c}
    \hline 
      Model  & w/o KD &  w/ KL-KD (T=1) & T=2  &T=4  & Adaptive-KD (Ours) \\ \hline
      MobileNetV3 $0.75\times$ & 73.3 &\bf 73.9 & 72.2 & 70.8 &\bf 73.9  \\ 
      MobileNetV3 $0.5\times$ & 69.6 & 69.8 & 65.4 & 63.6 & \bf 70.0 \\ \hline
    \end{tabular}
    \caption{
    Comparison to KL based KD with fixed teacher models on ImageNet.
    Here $T$ denotes the temperature used in classic KL based KD (see Appendix~\ref{app:kd}). 
    We use a MobileNetV3 $1.0\times$ as our teacher model, which yields $75.4\%$ top-1 validation accuracy on ImageNet. All MobileNetV3 student models are trained for 360 epochs with cosine learning rate decay.}
    \label{tab:mbv3}
    \vspace{1.5em}
    \begin{tabular}{c|cc|cc|c}
    \hline 
   \emph{Teacher}  & \multicolumn{2}{c|}{MobileNetV1 {~\small 1.0x}} & \multicolumn{2}{c|}{ MobileNetV2 {~\small 1.0x}}  & RegNetY \\
      \hline 
     \emph{Student}  &  ShuffleNet{~\small 0.5x} &  ShuffleNet{~\small 1.0x} & MobileNetV2{~\small 0.25x}  & MobileNetV2{~\small 0.5x} &  DeiT-tiny  \\
      \hline
      w/ KL-KD (T=1) & 60.3 & 69.3 &  54.4  & 65.3 & 74.6  \\
      Adaptive-KD (Ours) & \bf 61.1  & \bf 69.5  & \bf 55.0   & \bf  65.7   & \bf 75.2 \\
      \hline
    \end{tabular}
    \caption{Additional KD results on ImageNet. Our MobileNet V1 and V2 teacher has a top-1 accuracy of 73.2\% and 72.9\%, respectively. 
    All ShuffleNets~\citep{ma2018shufflenet} and MobileNetV2 models are trained for 120 epochs with standard random crop and resize data augmentation.
    For DeiT-tiny~\citep{touvron2020training}, we exactly follow the settings of DeiT for training and use a RegNetY~\citep{radosavovic2020designing} as the teacher model.}
    \label{tab:kd_standard}
\end{table*}

\subsection{Transfer learning}
\label{sec:exp_transfer}
Here we show that our AlphaNet models are not over-fitted on ImageNet and the knowledge learned on ImageNet could be transferred to other datasets as well. Specifically, we take our AlphaNet-A0 and AlphaNet-A6 models pretrained on ImageNet and fine-tune them on a number of transfer learning benchmarks.
We closely follow the training settings in EfficientNet~\citep{tan2019efficientnet} and GPipe~\citep{huang2018gpipe}. 
We use SGD with momentum of 0.9, label smoothing of 0.1 and dropout of 0.5. 
All models are fine-tuned for 150 epochs with batch size of 64. 
Following \citet{huang2018gpipe}, we search the best learning rate and weight decay on a hold-out subset (20\%) of the training data. 

\paragraph{Transfer learning results} 
We evaluated on five transfer learning benchmark datasets, including Oxford Flowers \citep{nilsback2008automated}, Oxford Pets \citep{parkhi2012cats}, Food-101~\citep{bossard2014food}, Stanford Cars~\citep{krause2013collecting} and Aircraft~\citep{maji2013fine}.
As we can see from Table~\ref{tab:transfer_learning}, 
our AlphaNet-A0 and  AlphaNet-A6 models lead to significant better transfer learning accuracy compared to those from EfficientNet-B0 and EfficientNet-B1 models.

\subsection{Additional results}
\label{sec:exp_ablation}

\paragraph{Robustness w.r.t. clipping factor $\pmb{\beta}$}
We follow the training and evaluation settings in section~\ref{sec:exp_supernet} and study the effect of $\beta$.
In Figure~\ref{fig:abla_alpha} (a), we group sub-networks according to their FLOPs, and report the relative top-1 accuracy improvements of the maximum top-1 accuracy of each FLOPs group over the result from the KL based KD baseline.   
As shown in Figure~\ref{fig:abla_alpha}(a),  our algorithm is robust to the choice of $\beta$. Our algorithm works with a large range of $\beta$, from 1 to 10, yielding consistent improvements over the classic KL based KD baseline. 
And our default setting $\beta=5$ achieves best performance on all FLOPs regimes evaluated.

\paragraph{Robustness w.r.t. $\pmb{\alpha}$}
We ablate the impact of both $\amin$ and $\amax$ under the same settings as in section~\ref{sec:exp_supernet}.
In this case,  we fix $\beta=5$.
We present our findings in Figure~ \ref{fig:abla_alpha}(b). 

Firstly,  we test with $\amin=-2, -1, 0$, with $\amax$ fixed as $1$. 
With a more negative $\alpha$ (e.g., $\amin=-2$), this defines a more difficult objective that brings optimization challenges. With a large $\amin$ (e.g., $\amin=0$), the resulting KD loss is less discriminative regarding {uncertainty over-estimation}. Overall, $\amin=-1$ achieves a good balance between optimization difficulty and over-estimation penalization, yielding the best performance.
Secondly, we vary $\amax$ from $0.5$ to $2$, with $\amin$ fixed as $-1$. 
Similarly, we find that large $\amax $ (e.g, $\amax=1$) yields the best performance.
Lastly, we set both $\amin=\amax=-1$. In this case, 
we still achieve better performance compared to the results of our KL based KD baseline, indicating the importance of penalizing over-estimation in training sub-networks. 
Also, our adaptive KD that regularizes on both over-estimation and over-estimation achieves better performance in general.

\paragraph{Improvement on single network training}
To further demonstrate the broader applicability of our method, 
we apply our Adaptive-KD to train 
a single neural network with a pretrained teacher model, 
as in convectional KD setup (See Appendix~\ref{app:kd}). 

Specifically, in Table~\ref{tab:mbv3}, 
we use a MobileNetV3 $1.0\times$ \citep{howard2019searching} as our teacher model and 
train MobileNetV3 $0.5\times$ and  $0.75\times$ as our student models.  
In Table~\ref{tab:kd_standard}, we provide additional comparisons for training ShuffleNets~\citep{ma2018shufflenet}, MobileNetv2 models~\citep{sandler2018mobilenetv2} and more recent vision transformers~\citep{touvron2020training}~\footnote{\url{https://github.com/facebookresearch/deit}} with a fixed temperature of $1.0$.

We summarize the top-1 validation accuracy on ImageNet from the models trained with different KD strategies in both Table~\ref{tab:mbv3} and Table~\ref{tab:kd_standard}. 
The student models trained via our method yield the best accuracy.

%% file: tex/conclusion.tex
\section{Related work}

\paragraph{Neural architecture search (NAS)}
NAS offers a powerful tool to automate the design of neural architectures for challenging machine learning tasks~\citep[e.g.,][]{fang2020densely, fu2021cm, moons2020distilling, li2020block, peng2020cream}.
Early NAS solutions usually build upon black-box optimization, e.g. reinforcement learning \citep[e.g.,][]{zoph2016neural}, Bayesian optimisation \citep[e.g.,][]{kandasamy2018neural}, evolutionary algorithms \citep[e.g.,][]{real2019regularized}. These methods find good networks but are extremely computationally expensive in practice.

More recent NAS approaches have adopted weight-sharing \citep{pham2018efficient} to improve search efficiency. 
Weight-sharing based approaches often frame NAS as a constrained optimization and solve with continuous relaxations~\citep[e.g.,][]{liu2018darts, cai2018proxylessnas}. 
However, these methods require to run NAS for each deployment consideration, 
 e.g. a specific latency constraint for a particular mobile device, 
the total search cost grows linearly with the number of deployment considerations  \citep{cai2019once}. 

To further alleviate the aforementioned limitations, 
one-shot supernet-based NAS~\citep[e.g.,][]{cai2019once, yu2020bignas, wang2020attentivenas} proposes to first jointly train 
 all candidate sub-networks specified in the weight-sharing graph such that all sub-networks reach good performance at the end of training; 
 then one can apply typical search algorithms, e.g., genetic search, to find a set of Pareto optimal networks for various deployment scenarios. Overall, one-shot supernet based methods provide a highly flexible and efficient NAS framework, yielding state-of-the-art empirical NAS performance on various challenging applications~\citep[e.g.,][]{cai2019once, wang2020hat}.
\paragraph{Knowledge Distillation} 
Our knowledge distillation forces the student model to mimic the predictions of the teacher model.
As shown in the literature, the features in intermediate layers of the teacher model can also be used as knowledge to supervise the training of the student model, 
notable examples include \citep[e.g.,]{romero2014fitnets,  huang2017like, ahn2019variational, jang2019learning, passalis2018learning, li2019hint}. 
Furthermore, correlations between different training examples (e.g. similarity) learned by the teacher model also provide rich information, which could be distilled to the student model \citep{park2019relational, yim2017gift}.
However, in our work, our KD involves training a large amount of sub-networks (students) with different architecture configurations, e.g., different network depth, channel width, etc. 
It is less clear on how to define a good matching in the latent feature space between the teacher supernet and student sub-networks in a consistent way. 
While our method offers a simple distillation mechanism that is easy to use in practice and in the meantime, leads to significant empirical improvements. 

\section{Conclusion}
In this work, we propose a method to improve the training of supernets with $\alpha$-divergence based knowledge distillation. 
By adaptively selecting an $\alpha$-divergence to optimize, 
our method simultaneously penalizes \emph{over-estimation} and \emph{under-estimation} in KD. 
Applying our method for neural architecture search, the searched AlphaNet models establish the new state-of-the-art accuracy vs. FLOPs trade-offs on the ImageNet dataset.

%% file: tex/appendix.tex
\newpage
\onecolumn
\appendix

\section{Weight-sharing NAS}
\label{app:one_shot_nas}

\begin{figure*}[h]
    \centering
    \includegraphics[width=0.6\textwidth]{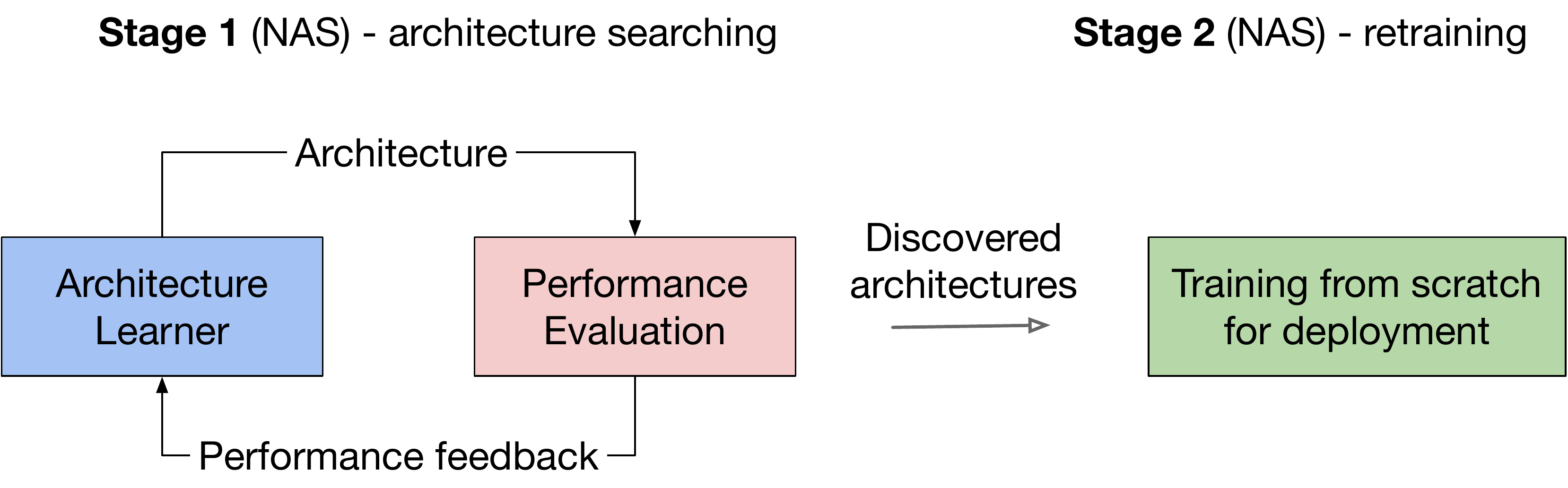}
    \caption{An overview of convectional NAS pipeline.}
    \label{fig:nas_sample}
\end{figure*}

Most  RL-based NAS~\citep[e.g.,][]{zoph2016neural} and differentiable NAS~\citep{liu2018darts, cai2018proxylessnas} 
consist of the following two stages as shown in Figure~\ref{fig:nas_sample}: 
\begin{enumerate}
\item[1)]  Stage 1 (architecture searching) - 
search potential architectures following a single resource constraint by using blackbox optimization techniques~\citep[e.g.,][]{zoph2016neural} or differentiable weight-sharing based approaches~\citep[e.g.,][]{liu2018darts, cai2018proxylessnas}; 
 \item[2)] Stage 2 (retraining) - retrain deep neural networks (DNNs) found in step 1) from scratch for best accuracy and final deployment.
\end{enumerate}

Though promising results have been demonstrated, these NAS methods usually suffer from the following disadvantages: 
1) need to re-do the NAS search for different hardware resource constraints; 
2) require training the selected candidate from scratch to achieve desirable accuracy; 
3) 1) especially for RL-based NAS that uses black-box optimization techniques, it requires training a large number of neural networks from scratch or on proxy tasks; 
These disadvantages significantly increase the computational cost of NAS and make the NAS search computationally expensive.

\begin{figure*}[h]
    \centering
    \includegraphics[width=0.65\textwidth]{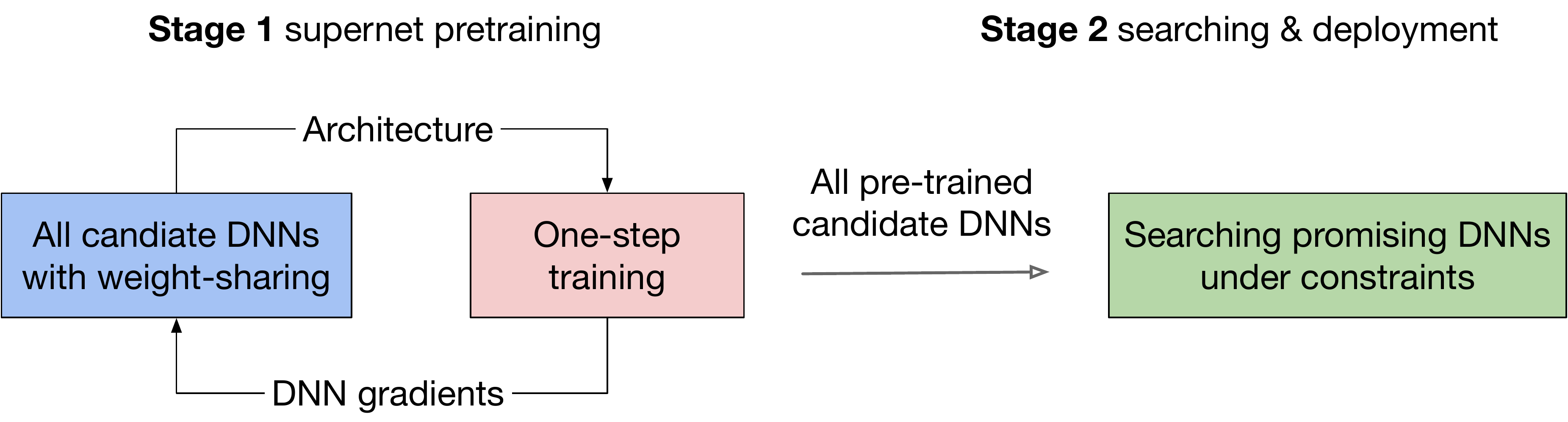}
    \caption{An overview of supernet-based weight-sharing NAS.}
    \label{fig:fastnas}
\end{figure*}

\paragraph{Supernet-based Weight-sharing NAS}
To alleviate the aforementioned issues, supernet-based  weight-sharing NAS transforms the previous NAS training and search procedures as follows; see Figure~\ref{fig:fastnas}.
\begin{enumerate}
\item[1)] Stage 1  (supernet pretraining): jointly optimize the supernet and all possible sub-networks specified in the search space, such that all searchable networks simultaneously achieve good performance at the end of the training phase.

\item[2)]  Stage 2 (searching \& deployment): 
After stage 1 training, all the sub-networks are optimized simultaneously. 
One could then use typical searching algorithms, like evolutionary algorithms, to search the best model of interest. The model weights of each sub-network are directly inherited from the pre-trained supernet without any further re-training or fine-tuning.
\end{enumerate}

Compared to RL-based NAS and differentiable NAS algorithms, the key advantages of the supernet-based weight-sharing NAS pipeline are: 1) one needs to only perform the computationally expensive supernet training for once. All sub-networks defined in the search space are ready to use after stage 1 is fully optimized. No retraining or fine-tuning is required;
2) all sub-networks of various model sizes are jointly optimized in stage 1, finding a set of Pareto optimal models that naturally supports various resource considerations. 

Notable examples of supernet-based weights-sharing NAS include
BigNAS~\citep{yu2020bignas}, OFA~\citep{cai2019once}, AttentiveNAS~\citep{wang2020attentivenas} and HAT~\citep{wang2020hat}.

\section{Weights-sharing NAS training settings}
\label{app:supernet}
We exactly follow the training settings in~\citet{wang2020attentivenas}~\footnote{\url{https://github.com/facebookresearch/AttentiveNAS}}. 
Specifically, 
we train our supernets for 360 epochs with cosine learning rate decay. 
We adopt SGD training on 64 GPUs. The mini-batch size is 32 per GPU. 
We use momeutm of 0.9, weight decay of $10^{-5}$, dropout of $0.2$, stochastic layer dropout of $0.2$. 
The base learning rate is set as 0.1 and is linearly scaled up for every 256 training samples. 
We use  AutoAugment \cite{cubuk2018autoaugment}  for data augmentation and set label smoothing coefficient to $0.1$.

We use the same search space provided in~\citet{wang2020attentivenas}, see Table~\ref{tab:fbnet_reduced_se}. 
Here Conv denotes regular convolutional layers and 
MBConv refers to inverted residual block proposed by~ \citet{sandler2018mobilenetv2}. 
We use swish activation. 
Channel width represents the number of output channels of the block. 
MBPool denotes the efficient last stage in~ \citet{howard2019searching}. 
SE represents the squeeze and excite layer ~\citep{hu2018squeeze}. 
\emph{Input resolution} denotes the candidate resolutions. 
To simplify the data loading procedure, 
we always pre-fetch training patches of a fixed size, e.g., 224x224 on ImageNet, and then rescale them to our target resolution with bicubic interpolation following \citep{yu2020bignas}.

\begin{table*}[ht]
    \centering
    \setlength\tabcolsep{10pt}
    \begin{tabular}{c|ccccc}
    \hline
     Block name & Channel width & Depth  & Kernel size  & Expansion ratio & SE \\ \hline
     Conv &  \{16, 24\} & - & 3 & - & - \\
     MBConv-1 &  \{16, 24\} &  \{1,2\} & \{3, 5\} & 1 & N\\
     MBConv-2 &  \{24, 32\} & \{3, 4, 5\} & \{3, 5\} &  \{4, 5, 6\} & N \\
     MBConv-3 &  \{32, 40\} & \{3, 4, 5, 6\} &\{3, 5\} & \{4, 5, 6\} & Y\\
     MBConv-4 &  \{64, 72\} & \{3, 4, 5, 6\} &\{3, 5\} & \{4, 5, 6\} & N\\
     MBConv-5 &  \{112,128\} & \{3, 4, 5, 6, 7, 8\} & \{3, 5\} & \{4, 5, 6\} & Y\\
     MBConv-6 &  \{192, 200, 208, 216\} & \{3, 4, 5, 6, 7, 8\}  &\{3, 5\} & 6 & Y\\
     MBConv-7 &  \{216, 224\} & \{1, 2\}  &\{3, 5\} & 6 & Y\\
     MBPool  & \{1792, 1984\} & - & 1 & 6 & - \\
     \hline
     Input resolution &   \multicolumn{4} {c}{\{192, 224, 256, 288\}} \\
    \hline
    \end{tabular}
    \caption{An illustration of our search space. Every row denotes a block group. 
    }
    \label{tab:fbnet_reduced_se}
\end{table*}

\section{Knowledge distillation}
\label{app:kd}
Consider the image classification task over a set of classes $[m]:= \{1,\cdots, m\}$,
where we have a collection of training images and one-hot labels 
$\mathcal{D}^{train}=\{(x, y)\}$ with $(x, y)\in \mathcal{X}\times \mathcal{Y}$
and $y\in\{0, 1\}^m$. 
We are interested in designing a deep neural network $\ss(x; \theta): \mathcal{X} \rightarrow \mathcal{Y}$ that captures the relationship between $x$ and $y$. 
Here $\theta$ is the network parameters of interest. 

KD provides an effective way to train $\ss$ by distilling knowledge from a teacher model in addition to the one-hot labels. 
The teacher network is often a relative larger network with better performance. 
Specifically, let $\tt$ be the teacher network, KD enforces $\ss$ to mimic the output of $\tt$ by minimizing the closeness between $\ss$ and $\tt$, which is often specified by the KL divergence $\D_{\KL}(\tt\pp\ss)$, 
yielding the following loss function, 
\begin{align}
\L&(\theta) = (1-\beta)\L_{\erm}(\theta) + 
\beta  \L_{\KD}(\theta),~~~\text{with} \nonumber \\
&\L_{\erm}(\theta) = \E_{(x, y)\sim \mathcal{D}^{train}} \bigg[ \L(y, \ss(x; \theta)) \bigg], \nonumber \\
&\L_{\KD}(\theta)= \E_{x\sim \mathcal{D}^{train}}\bigg[ \D_{\KL}(\tt(x)\pp\ss(x;\theta)) \bigg]. \label{eq:kd}
\end{align}
Here $\L(\cdot)$ represents the empirical loss,  e.g., the typical cross entropy loss $\L(y, \ss(x;\theta))=\sum_{i=1}^m -y_i \log \ss_i$ with $\ss_i$ be the $i$-class probability produced by $\ss$. And $\D_{\KL}(\tt\pp\ss) = \E_{\tt}[\log (\tt/\ss)]$. Furthermore, $\beta \in [0,1]$ is the distilling weight that balances of the empirical loss and KD loss.

One could also apply a temperature $T$ to soften (or sharpen) the outputs the teacher model and the student model in KD.  More precisely, given an input $x$, we assume $z_i^{\tt}(x)$ and $z_i^{\ss}(x)$ the logit for the $i$-th class produced by $\tt$ and $\ss$, respectively. Then the corresponding predictions of $\tt$ and $\ss$ after temperature scaling are as follows,  
\begin{align*}
\tt_i(x; T) = \mathrm{softmax}(z_i^\tt; T),~~  \ss_i(x; T) = \mathrm{softmax}(z_i^\ss; T),
\end{align*}
with $\mathrm{softmax}(z_i; T) = {\exp(z_i/T)}/ {\sum_i \exp(z_i/T)}$. 
In this way, the previous KD objective~\eqref{eq:kd} 
is now adapted to the following, 
\begin{align}
\L(\theta; T) &= (1-\beta) \L_{\erm}(\theta) +\beta T^2 \L_{\KD}(\theta; T), ~~ \text{with}\nonumber \\  
\L_{\KD}&(\theta; T)= \E_{x}\bigg[ D_{\KL}(\tt(x; T)\pp\ss(x; T, \theta)) \bigg].   \label{eq:kd_t}
\end{align} 
Here $T^2$ is introduced to ensure the gradients from the KD loss is at the same scale w.r.t the gradients from the empirical loss, see~\citep[e.g.,][]{hinton2015distilling}. 
We set $\beta=0.9$ as default.

\newpage
\section{Additional results on ablation studies}
Following the settings in section~\ref{sec:exp_supernet}, 
we provide further analyses on the performance of sub-networks learned under 
different $\alpha$ and $\beta$ settings. 

\begin{figure*}[ht]
\begin{tabular}{c}
\raisebox{3em}{\rotatebox{90}{\small Validation Accuracy}}
\includegraphics[width=0.93\textwidth]{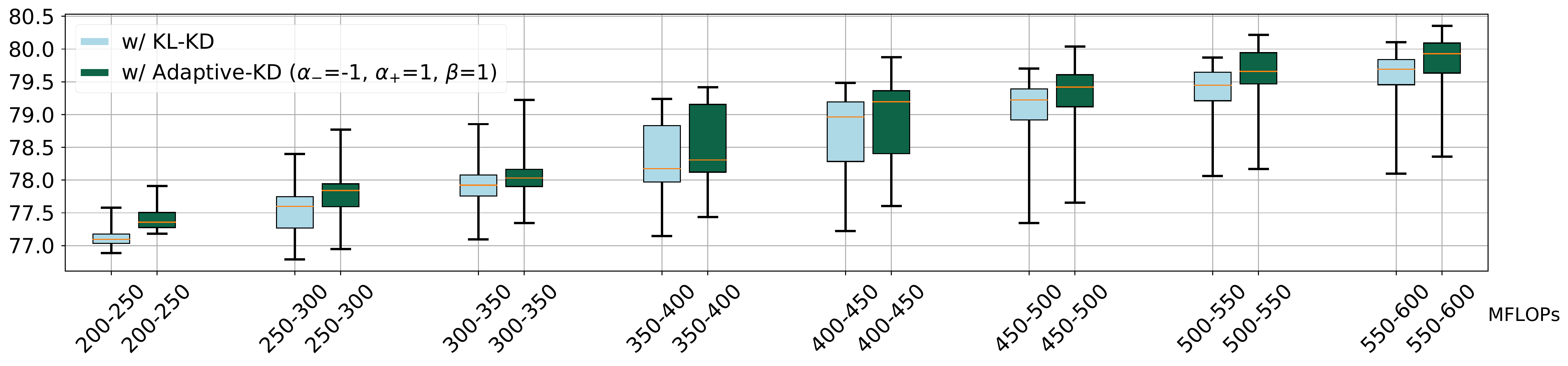}  \\
\raisebox{3em}{\rotatebox{90}{\small Validation Accuracy}}
\includegraphics[width=0.93\textwidth]{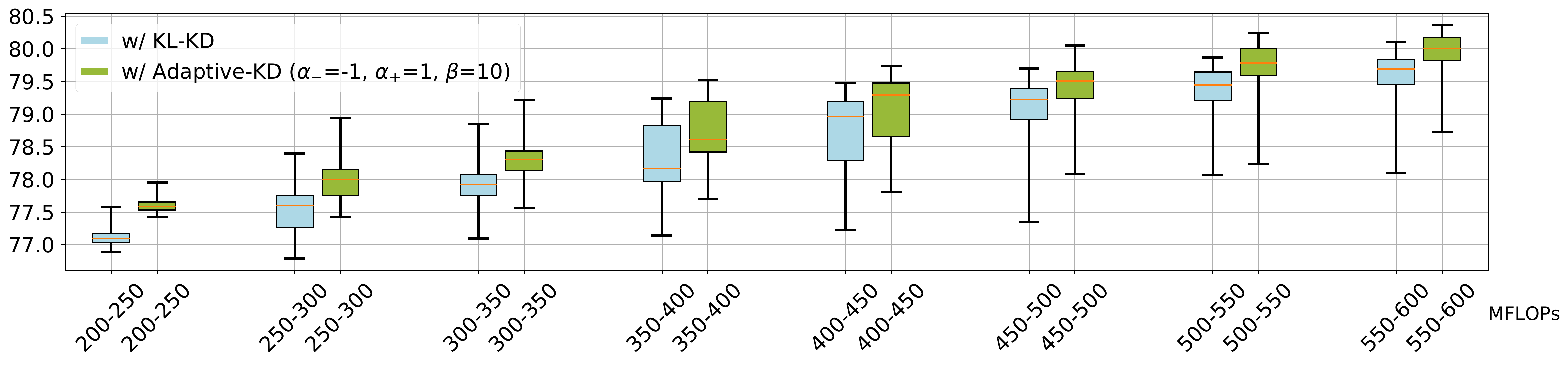}  \\
\raisebox{3em}{\rotatebox{90}{\small Validation Accuracy}}
\includegraphics[width=0.93\textwidth]{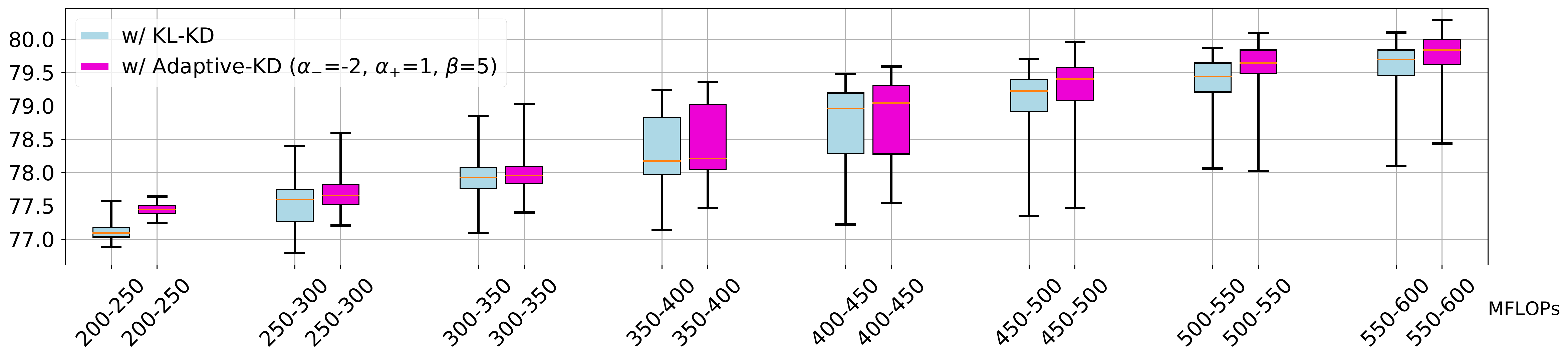}  \\
\raisebox{3em}{\rotatebox{90}{\small Validation Accuracy}}
\includegraphics[width=0.93\textwidth]{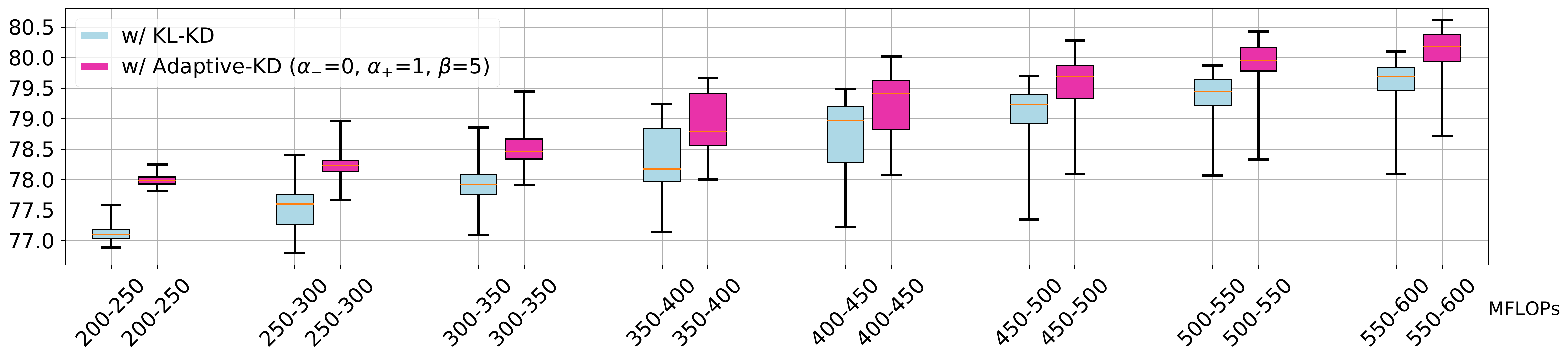}  \\
\end{tabular}
\label{fig:app_abalate_supernet_boxplot}
\caption{Additional results on ablation studies. Each box plot shows the performance of sampled sub-networks within each FLOPs regime. From bottom to top, each horizontal bar represents the minimum accuracy, the first quartile, the median, the third quartile and the maximum accuracy, respectively. }
\end{figure*}

\begin{figure*}[ht]
\begin{tabular}{c}
\raisebox{3em}{\rotatebox{90}{\small Validation Accuracy}}
\includegraphics[width=0.93\textwidth]{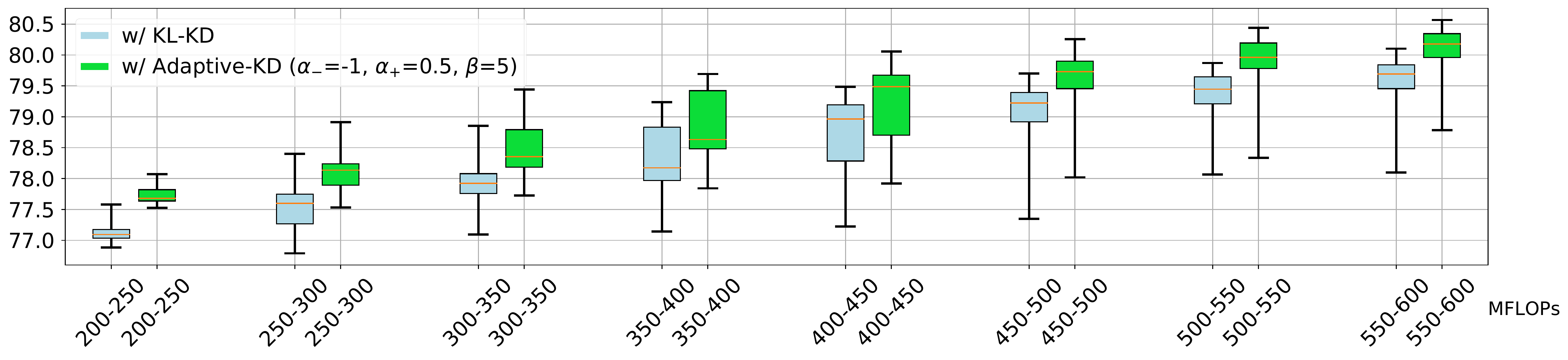}  \\
\raisebox{3em}{\rotatebox{90}{\small Validation Accuracy}}
\includegraphics[width=0.93\textwidth]{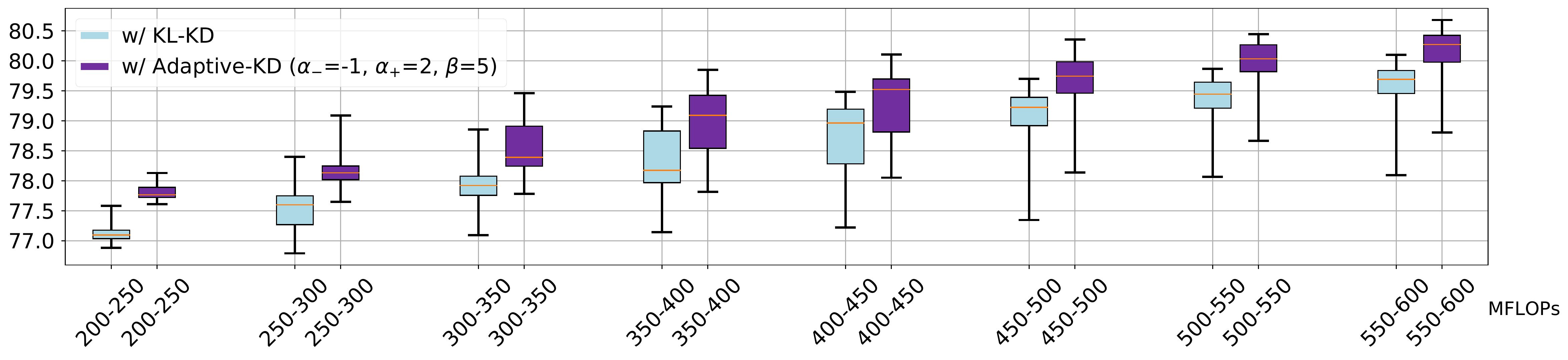}  \\
\raisebox{3em}{\rotatebox{90}{\small Validation Accuracy}}
\includegraphics[width=0.93\textwidth]{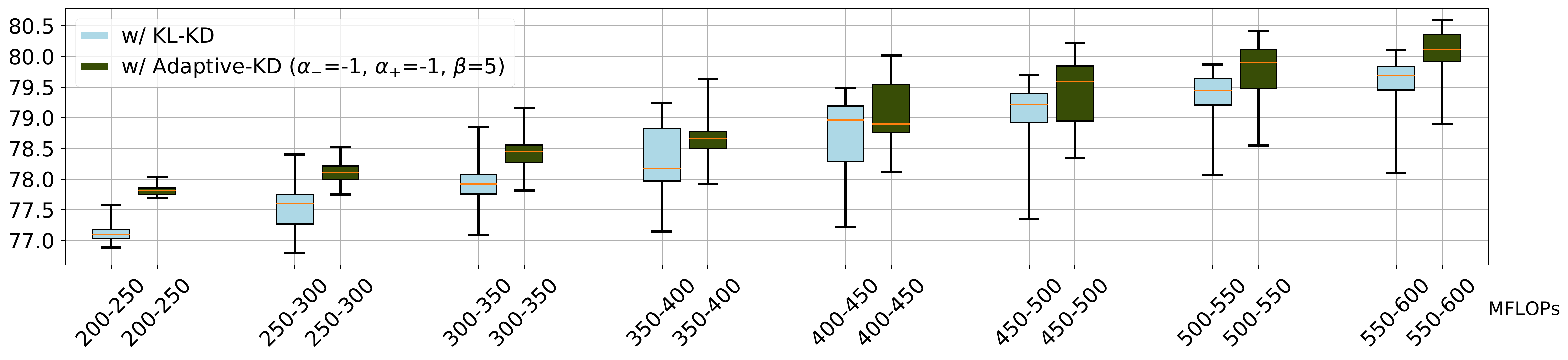}  \\
\end{tabular}
\label{fig:app_abalate_supernet_boxplot_cont}
\caption{Additional results on ablation studies. Each box plot shows the performance of sampled sub-networks within each FLOPs regime. From bottom to top, each horizontal bar represents the minimum accuracy, the first quartile, the median, the third quartile and the maximum accuracy, respectively. }
\end{figure*}